\documentclass[journal]{IEEEtran}
\usepackage{cite}

\ifCLASSINFOpdf
   \usepackage[pdftex]{graphicx}
\else
   \usepackage[dvips]{graphicx}
\fi
\usepackage{balance}

\usepackage{amsmath}
\usepackage{algorithmic}
\usepackage{algorithm}
\usepackage{amsmath}

\usepackage{array}
\usepackage{fixltx2e}

\usepackage{stfloats}
\ifCLASSOPTIONcaptionsoff
 \usepackage[nomarkers]{endfloat}
\let\MYoriglatexcaption\caption
\renewcommand{\caption}[2][\relax]{\MYoriglatexcaption[#2]{#2}}
\fi
\usepackage{url}

\usepackage{subcaption}
\usepackage{tikz}
\usepackage{tikzsymbols}
\usepackage{verbatim}
 \usepackage{pifont} 
\usetikzlibrary{shapes.geometric, arrows.meta, positioning}
\tikzstyle{startstop} = [rectangle, rounded corners, minimum width=3cm, minimum height=1cm,text centered, draw=black, fill=gray!20]
\tikzstyle{process} = [rectangle, minimum width=3cm, minimum height=1cm, text centered, draw=black, fill=blue!10]
\tikzstyle{decision} = [diamond, aspect=2, text centered, draw=black, fill=yellow!20]
\tikzstyle{arrow} = [thick, ->, >=stealth]

\begin{document}
%
\title{Beyond Real versus Fake \\Towards Intent-Aware Video Analysis}
%
%
%

\author{Saurabh~Atreya,
        Nabyl~Quignon,
        Baptiste~Chopin,
        Abhijit~Das,~\IEEEmembership{Senior Member,~IEEE}, 
        Antitza~Dantcheva,~\IEEEmembership{Senior Member,~IEEE}

\thanks{S. Atreya and A. Das are with BITS~Pilani~Hyderabad, India, e-mail: abhijit.das@hyderabad.bits-pilani.ac.in}
\thanks{N. Quignon and A. Dantcheva are with the Inria Center at Université Côte d'Azur, France.}
\thanks{B. Chopin is with the Inria Center at Université Côte d'Azur, France and da/sec – Biometrics and Security Research Group, Hochschule Darmstadt, Germany}
\thanks{Manuscript received May 31, 2025.}}


%
%

\markboth{Journal of \LaTeX\ Class Files, May~2025}%
{Shell \MakeLowercase{\textit{et al.}}: Bare Demo of IEEEtran.cls for IEEE Journals}
%



\maketitle

\begin{abstract}
The rapid advancement of generative models has led to increasingly realistic deepfake videos, posing significant societal and security risks. While existing detection methods focus on distinguishing real from fake videos, such approaches fail to address a fundamental question: What is the intent behind a manipulated video? Towards addressing this question, we introduce IntentHQ: a new benchmark for human-centered intent analysis, shifting the paradigm from authenticity verification to contextual understanding of videos. IntentHQ consists of 5168 videos that have been meticulously collected and annotated with 23 fine-grained intent-categories, including 'Financial fraud', 'Indirect marketing', 'Political propaganda', as well as 'Fear mongering'.
We perform intent recognition with supervised and self-supervised multi-modality models that integrate spatio-temporal video features, audio processing, and text analysis to infer  underlying motivations and goals behind videos.
Our  proposed model is streamlined to differentiate between a wide range of intent-categories.
{The IntentHQ dataset is available at \url{https://github.com/atrey-a/intent-hq}.} 
\end{abstract}


%
\IEEEpeerreviewmaketitle

\section{Introduction}

In the past decade, we have witnessed remarkable progress in generative models, successfully creating unprecedented levels of \textit{increasingly realistic human videos}. While such technology bears highly \textit{exciting perspectives} for entertainment, marketing, as well as education, it poses an immense challenge to societal trust, public safety, and digital security. In particular, humans in generated videos can be depicted in actions that have not taken place, commonly referred to as deepfakes. Such deepfakes have already been weaponized for \textit{nefarious purposes}, including disinformation campaigns\footnote{https://www.bbc.com/news/technology-60780142}, financial fraud\footnote{
  https://edition.cnn.com/2024/05/16/tech/arup-deepfake-scam-loss-hong-kong-intl-hnk/index.html}, and non-consensual sexual personal harassment\footnote{https://www.nytimes.com/2024/04/08/technology/deepfake-ai-nudes-westfield-high-school.html}. Even more worrying is the fact that – while in the past, video forgery was associated with a slow, painstaking process usually reserved for experts – currently, deepfake-related manipulation technologies are streamlined to be used by anyone with the intent to manipulate reality in real time based on a single image of someone. Such deepfake techniques are now widespread via phone applications such as DeepFaceLab, FakeApp, and Zao, and related results can be rapidly disseminated via social media.

To date, deepfakes are able to mislead AI algorithms, as well as humans \cite{porcile2024finding,moreira2024synthetic}. Despite increased awareness of the threat, the cat-and-mouse game between generation and detection is currently led by \textit{generation} - a research area fueled by the fierce competition of \textit{data- and compute-rich industrials} such as OpenAI, Google, NVIDIA, Microsoft, and Meta. In comparison, the \textit{detection} side remains \textit{limited} in scope and resources \cite{lin2024detecting}. 

In addition, with forecasts indicating that by the end of $2025$, $90\%$ of online images and videos will be AI-generated\footnote{https://finance.yahoo.com/news/90-of-online-content-could-be-generated-by-ai-by-2025-expert-says-201023872.html}, the \textit{distinction between real and fake is increasingly blurring}, rendering the current binary classification approach to deepfake detection (DD) inadequate. What we propose here is to shift focus from such binary classification to \textbf{intent characterization in videos}. Specifically, rather than asking whether a video is real or generated, the question will shift to whether it is malicious or benign. This focus is further motivated by the current evolution of AI agents, targeted to be autonomously reasoning, acting, and even collaborating with other agents\footnote{https://openai.com/index/altera/}. However, even before their full sophistication is achieved, concerns are emerging about possible hidden agendas \cite{chaudhary2024large,meinke2024frontier}, and how an entire industry may emerge that seeks to covertly influence decision-making\footnote{https://www.cam.ac.uk/research/news/coming-ai-driven-economy-will-sell-your-decisions-before-you-take-them-researchers-warn}. The above makes it imperative to develop methods for recognizing intent in videos — a critical objective of this work.

Motivated by the above, our contributions include the following. 

\begin{itemize}
    \item We formalize the new task of \textbf{intent recognition} in videos, which goes beyond conventional content-based video understanding. Specifically, we focus on identifying the underlying \textit{purpose or motivation} of a video, thereby enabling a shift from authenticity-based classification to intent-aware interpretation. Intent recognition is a complementary task to deepfake detection with the aim of informing users about the content they consume online.

    \item We introduce \textbf{IntentHQ}, a new dataset that we have curated specifically for intent recognition in human-centric videos. IntentHQ 
    comprises $5168$ manually annotated YouTube videos across $23$ intent categories. 

    \item We benchmark intent recognition by \textbf{existing pretrained single-modality models} (\textit{e.g.,} video-, audio-, as well as text-encoders), and subsequently train \textbf{combined multi-modal models} in a supervised setting. These results serve as the first strong supervised baseline for IntentHQ.

    \item Finally, we propose a novel \textbf{self-supervised learning approach} trained on the IntentHQ dataset. Our method leverages cross-modal contrastive learning, aimed at aligning video, audio, and textual modalities, without requiring manual intent labels. This self-supervised model outperforms the initial benchmark. 
\end{itemize}

\begin{figure*}[ht]
    \centering
    \includegraphics[width=\linewidth]{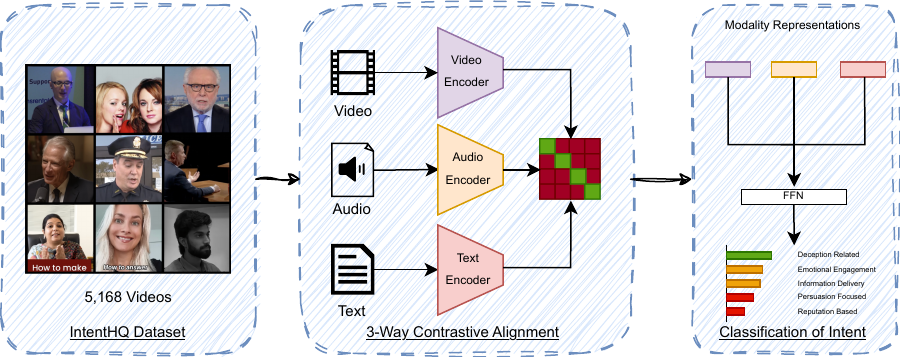}
    \caption{\textbf{Three-Way Contrastive Alignment Pipeline.} Overview of the proposed training methodology. The augmented dataset is encoded using modality-specific encoders (CLIP for video, WavLM for audio, CLIP Text for text), projected into a shared space, and aligned through a three-way contrastive loss. The pretrained encoders are then fine-tuned using a supervised MLP classifier for intent prediction.}
    \label{fig:three-way-contrastive}
\end{figure*}

\section{Related Work}


\subsection{Video Understanding}
Video understanding aims at extracting video-features, in order to analyze and categorize these for tasks such as action recognition \cite{siddiqui2024dvanet,xiong2024modality}, object detection \cite{hui2024endow,mahmud2024ssvod}, scene understanding \cite{ristea2024self,wu2024weakly} as well as dense video captioning \cite{kim2024you,zhou2024streaming}. Architectures of such methods evolved from 2D CNNs \cite{shao2016slicing} to 3D CNNs \cite{wu2019long,feichtenhofer2020x3d}, in order to capture temporal information. Further, Two-Stream networks \cite{simonyan2014two} and Temporal segment networks (TSNs) \cite{wang2016temporal} have been introduced to improve the temporal modeling. More recently, Transformer and other attention-based models have been adopted to capture long range dependencies \cite{li2022uniformer,liu2022video,wang2022internvideo}. Current state of the art (SoTA) for video understanding involves Mamba and Vision Transformer \cite{li2024mamba,park2024videomamba,reilly2024just}. Progress in video understanding is fueled by increased volumes of training datasets  \cite{soomro2012ucf101,caba2015activitynet,liu2022fineaction,miech2019howto100m,yu2019activitynet}. However, above methods and datasets are designed to only extract visual information from videos, ignoring the underlying intent behind such content. 

\subsection{Multimodal Language Models}
Recent advances related to Large Language Models (LLMs) \cite{radford2018improving,achiam2023gpt} and Large Vision-Language Models (LVMs)  \cite{dong2025internlm,ren2024timechat} led to the development of Multimodal Language Models (MLMs) that have the ability to process and understand multiple types of data jointly, such as text, images, audio, and video \cite{wu2025visionllm,peng2023kosmos}. MLMs can also be seen as an extension of video understanding models, as MLMs analyze videos  considering both, visual content, as well as accompanying audio. Flamingo \cite{alayrac2022flamingo} was the first MLM and was followed by a myriad of other models, each designed for specific tasks using various modalities. Some MLMs focus only on multimodal understanding using text and image \cite{liu2023visual}, video \cite{ma2024dolphins}, audio \cite{tang2023salmonn} 3D data \cite{xu2024pointllm} or multiple types of inputs \cite{yang2023mm,moon2024anymal,chen2023x}. Other MLMs also allow for multimodal content generation such as images \cite{lai2024lisa}, videos \cite{jin2024video}, audio speech \cite{zhang2023speechgpt} or multiple types at the same time \cite{wu2024next,wang2024tool,liu2024controlllm}. MLMs generally comprise of 
modules, each encoding a modality (\textit{e.g.}, CLIP \cite{pmlr-v139-radford21a} for text, HuBERT \cite{hsu2021hubert} for audio), an LLM backbone (\textit{e.g.}, LLaMA \cite{llama3modelcard}, Qwen \cite{yang2024qwen2}), and in certain cases generative modules (\textit{e.g.}, Stable diffusion \cite{rombach2022high} for image, AudioLDM \cite{liu2024audioldm} for audio).
However, while these models excel at capturing visual information and information related to  other modalities, to the best of our knowledge, they are not able to predict intent in videos.  
This can be attributed in part to lack of intent-related data, which is 
only reliably annotated by humans, limiting the collection of 
large scale datasets. Towards addressing this limitation, we here propose a new dataset and  benchmarking frameworks, trained for intent characterization in videos.

\subsection{Deepfake Detection}
The literature on detecting deepfake \textit{images} is extensive, whereas research on detecting deepfake \textit{videos} has received less attention, as revisited in overview articles \cite{wang2024deepfake,tolosana2020deepfakes}.

\noindent\textit{Artifact-based techniques} extract low-level to high-level features ranging from frequency \cite{cozzolino2021spoc} to blending \cite{shiohara2022detecting}, produced by generation models. Such methods are vulnerable to compression artifacts and adversarial attacks. \\
\textit{Learning-based Techniques} represent the majority of deepfake detection methods, utilizing data-driven learning methods that exploit 
numerous cues including eye blinking, head pose and facial behavior. At the same time, a large number of prominent network structures have been employed such as C3D, LSTM, RNN and most recently Video Transformer \cite{zhao2023istvt,kaddar2024deepfake}. \\
In contrast, another line of research proposes \textit{active deepfake detection} that incorporates watermarks and blockchain signatures into images, in order to ensure the authenticity of media. 

Currently, efforts in detection predominantly rely heavily on supervised learning, where algorithms infer binary labels (\textit{e.g.,} "real" or "fake") for new data based on annotated training datasets. These methods face significant limitations, struggling with data outside their training distribution and offering little to no transparency in their decision-making processes.

Post-processing and filters on off-the-shelf telephones challenge the definition and separability of real and deepfake videos. Further, the increasing amount of generated videos will inevitably shift the question of whether a video is real or generated to whether it is malicious, which we introduce here.

\subsection{Intent Recognition in Natural Language}

Intent recognition has been a key objective in the field of Natural Language Processing (NLP), particularly within dialogue systems, question answering, and virtual assistant frameworks. The goal in such settings is to determine the underlying purpose behind a user’s utterance, such as whether a query expresses a request, command, or expression of sentiment. Early approaches to intent recognition relied on rule-based systems or shallow classifiers using hand-crafted features \cite{tur2011spoken}. With the advent of deep learning, these were replaced by more robust methods such as RNN-based sequence models, attention mechanisms, and transformer architectures \cite{chen2019bertintent,zhang2020task}.

More recent intent recognition frameworks employ large pre-trained language models such as BERT \cite{devlin2018bert}, RoBERTa \cite{liu2019roberta}, or GPT variants, which are fine-tuned on task-specific data. These models demonstrate strong performance on intent classification tasks across datasets such as ATIS, SNIPS, and Banking77 \cite{casanueva2020efficient}, with the added benefit of generalizability and few-shot capabilities. Similarly, text based fake news detection methods \cite{10477989,10568915} aim to recognize fake news in written content also using large pre-trained language models or other deep networks such as CNN or LSTM.

Deviating from the above, we propose a multimodal approach, incorporating video, audio, as well as text. We note that while NLP datasets contain well-structured inputs, videos are inherently unstructured. Therefore, existing NLP intent recognition approaches cannot be directly applied to videos, particularly those involving generated or potentially misleading content.


\section{Proposed IntentHQ Dataset}


We have curated a new dataset, referred to as IntentHQ, targeted for intent recognition. To the best of our knowledge, this is the first dataset of its kind. Specifically, we have curated 
$5168$ videos, 
sourced from YouTube. The dataset contains high quality, human-centric videos, manually categorized into $23$ fine-grained intent classes. IntentHQ provides video, audio, as well as speech transcript - obtained with the Whisper-Large \cite{radford2023robust} speech to text model.

\subsection{Intent Categories}
\label{sec:class_list}

\begin{figure*}[!h]
    \centering
    \begin{tabular}{ccccc}
        \includegraphics[width=0.18\linewidth, height=0.18\linewidth, keepaspectratio=FALSE]{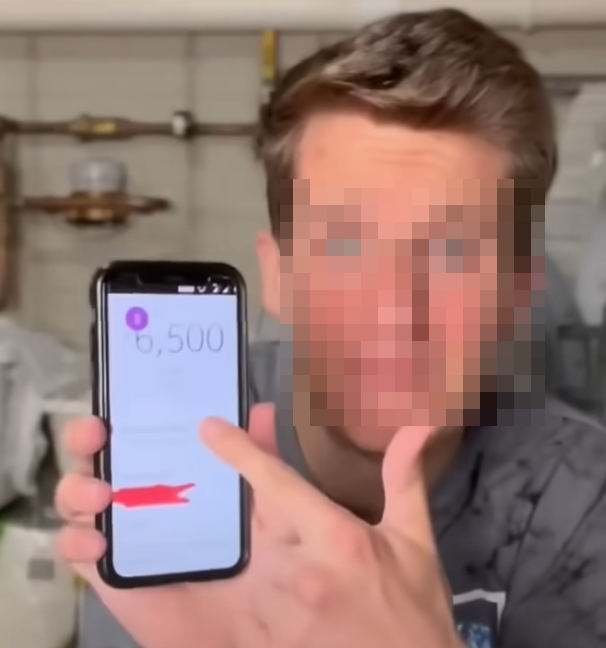} &
        \includegraphics[width=0.18\linewidth, height=0.18\linewidth, keepaspectratio=FALSE]{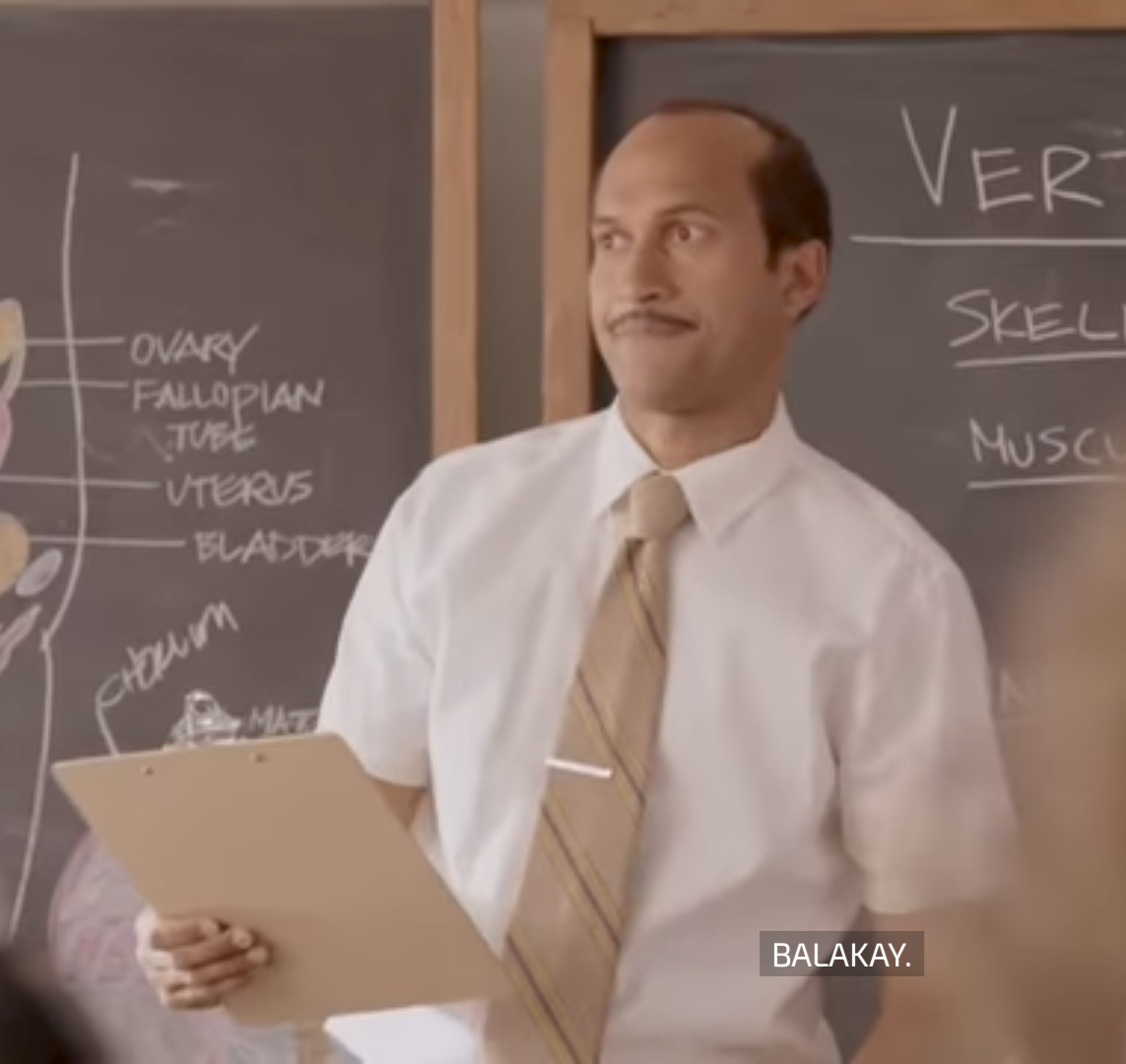} &
        \includegraphics[width=0.18\linewidth, height=0.18\linewidth, keepaspectratio=FALSE]{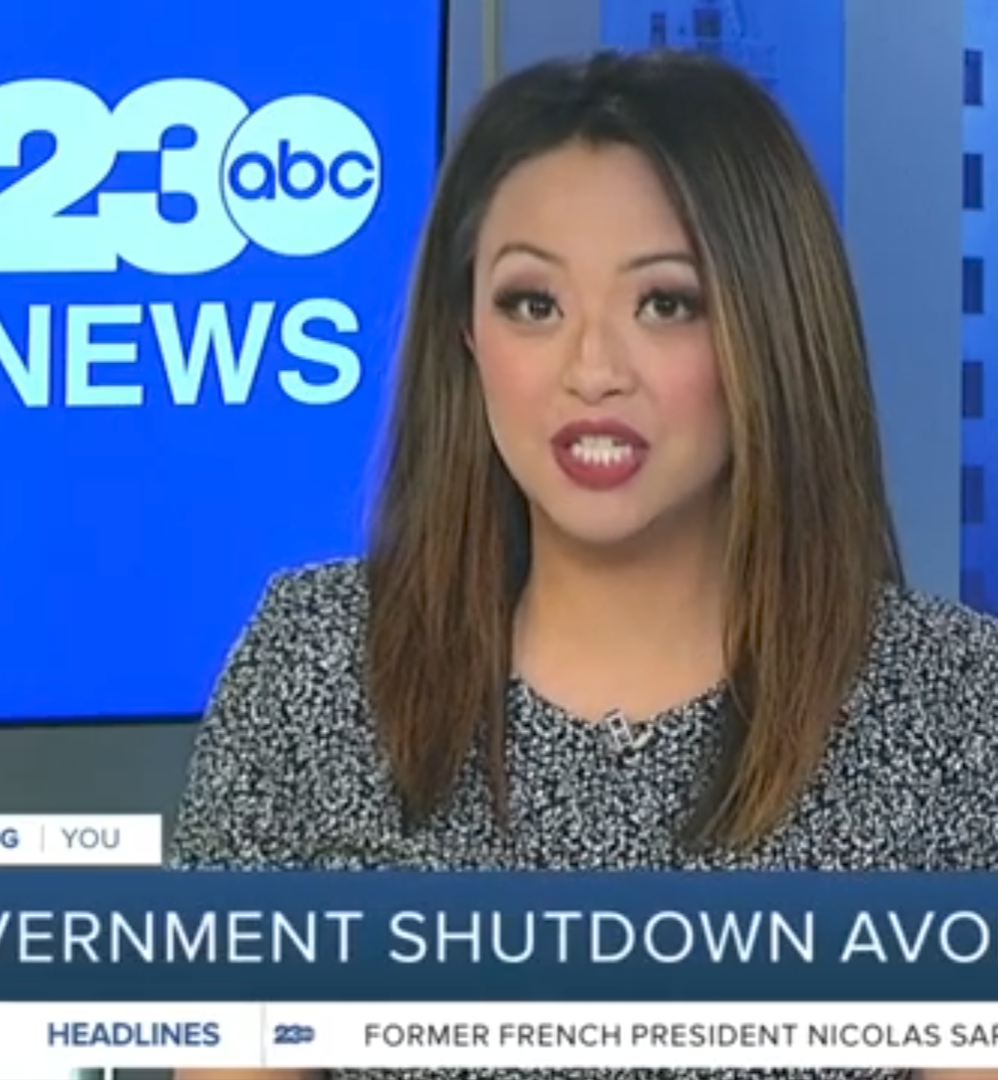} &
        \includegraphics[width=0.18\linewidth, height=0.18\linewidth, keepaspectratio=FALSE]{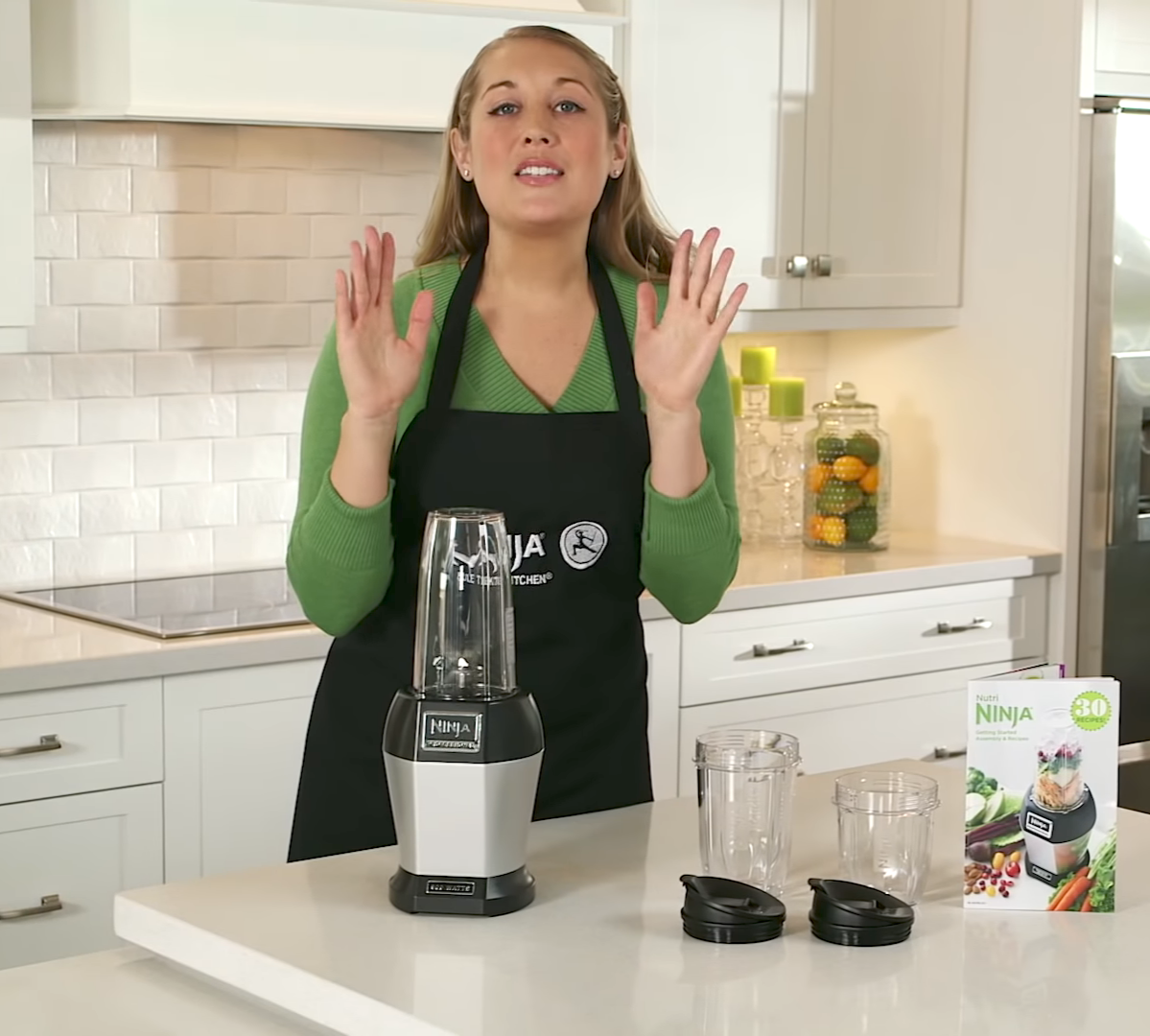} &
        \includegraphics[width=0.18\linewidth, height=0.18\linewidth, keepaspectratio=FALSE]{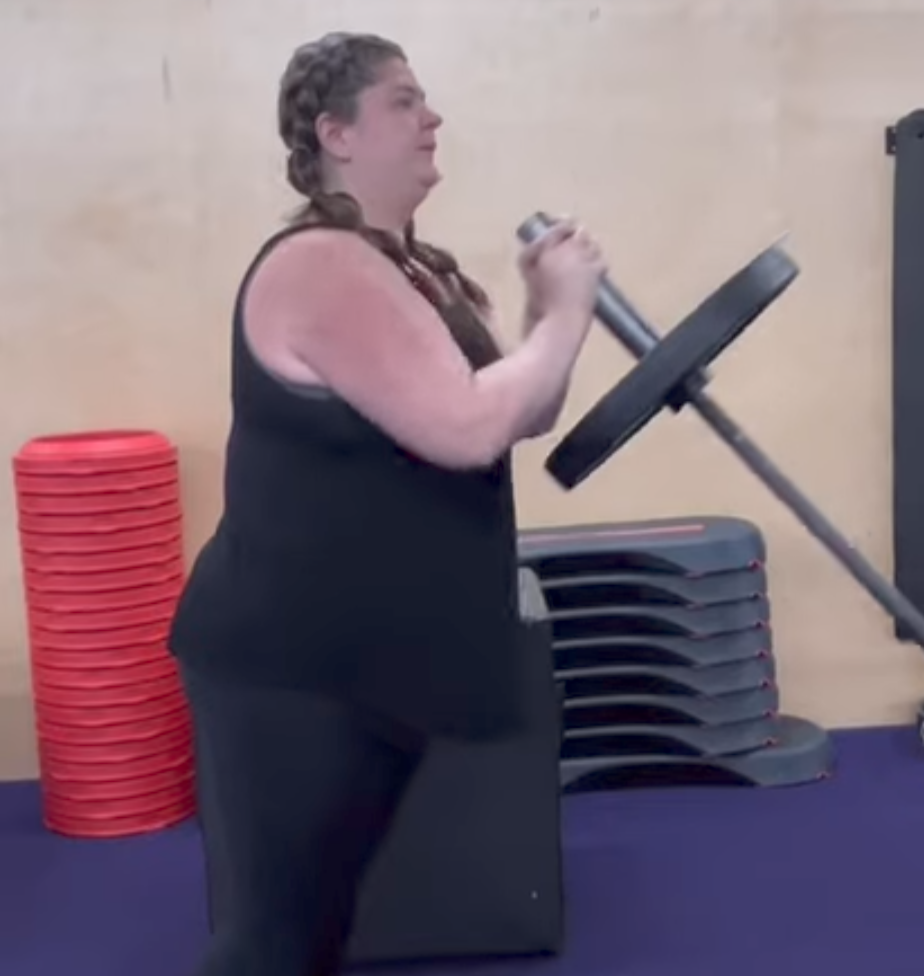} \\
        \scriptsize I. Deception & \scriptsize II. Emotional Engagement &  \scriptsize III. Information Delivery & \scriptsize IV. Persuasion Focused & \scriptsize V. Reputation
    \end{tabular}
    \caption{\textbf{Samples from our IntentHQ dataset.} IntentHQ contains $5$ broad categories further divided into $23$ total intent classes. The images above represent examples from each broad category.
    }
    \label{fig:sample_images}
\end{figure*}

We have identified $23$ intent categories spanning a wide spectrum of possible intentions. We group our $23$ sub-categories into $5$ main categories, pertained to I. Deception, II. Emotion, III. Information, IV. Persuasion, as well as V. Reputation, as shown in Figure \ref{fig:sample_images}.
The choice of categories ensures a diverse dataset that represents distinct aspects of human communication and intent \cite{talevich2017toward}, capturing a wide range of human motives. This diversity is crucial for training models that can generalize well across different scenarios and contexts. We proceed to elaborate on the categories.

\begin{figure*}[ht]
    \centering
    \includegraphics[width=\linewidth]{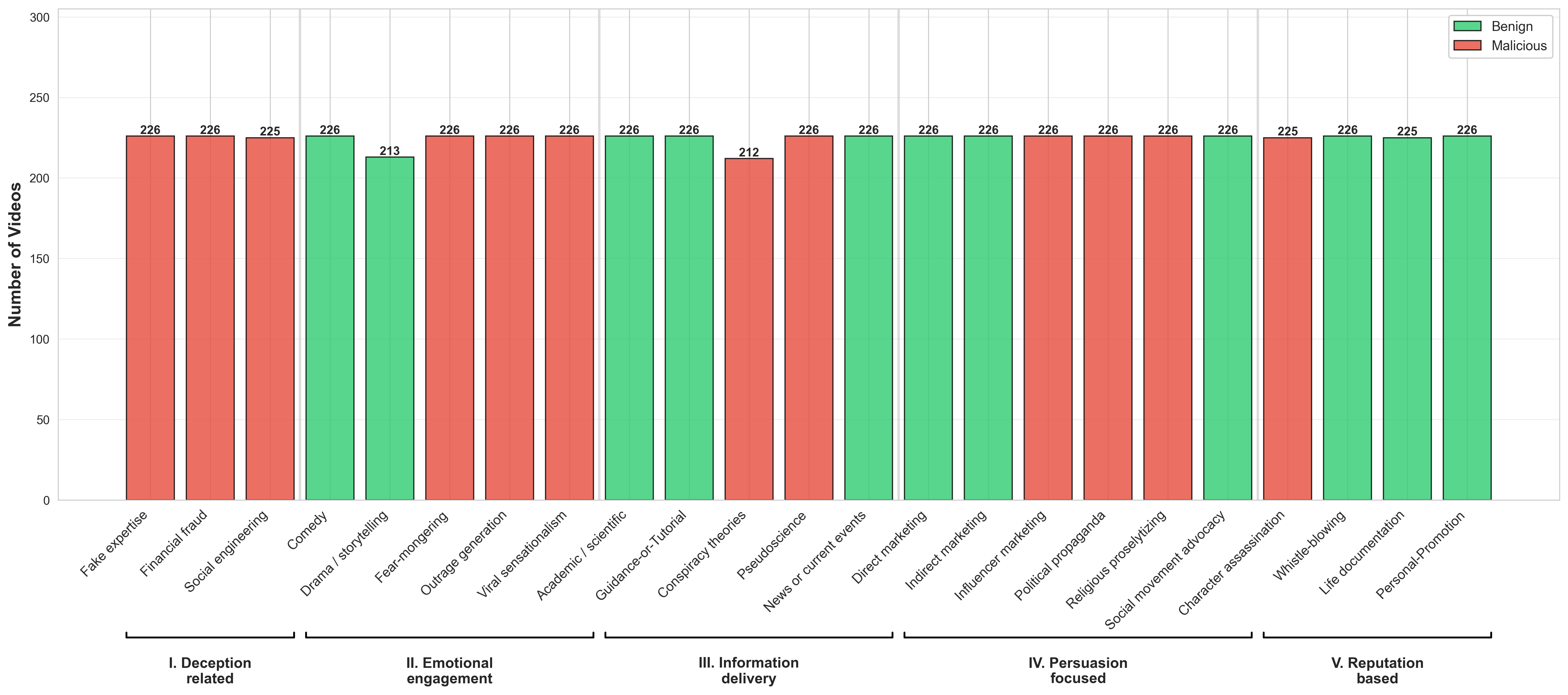}
    \caption{\textbf{Class Distribution within IntentHQ.} Overview of the number of videos within each class of IntentHQ, denoting the grouping of the larger main categories as well as showing whether classes are considered benign or malicious.}
    \label{fig:class_distribution}
\end{figure*}

\noindent{\textbf{I. Deception related}} videos include the sub-categories \textbf{1. Fake expertise}, where individuals attempt to deceive others by providing incorrect advice, while posing as experts on a subject. In \textbf{2. Financial fraud} videos, individuals encourage others to invest in fraudulent financial operations with the intent to steal their funds. \textbf{3. Social engineering} videos are designed to manipulate viewers through calculated deceptive tactics, in order to influence their behaviour or beliefs.

\noindent{\textbf{II. Emotional engagement}} videos contains \textbf{4. Comedy} which aims to elicit laughter in viewers, and \textbf{5. Drama / storytelling}, where narratives are crafted to elicit empathy. Next, \textbf{6. Fear-mongering} videos are designed to instill fear in people over specific subjects. Other emotion-focused content includes \textbf{7. Outrage generation} and \textbf{8. Viral sensationalism}.

\noindent{\textbf{III. Information delivery}} videos comprise \textbf{9. Academic or scientific} videos such as lectures or scientific popularization, as well as \textbf{10. Guidance-or-Tutorial} content, like tutorials, aiming to share theoretical or practical knowledge, respectively. Videos of \textbf{11. Conspiracy theories} discuss and promote unverified theories that attempt to explain historical events, usually through the action of powerful groups acting in secret. This category also includes \textbf{12. Pseudoscience} videos, which promote false or misleading information related to science. In contrast, \textbf{13. News or current event} content provides real and verifiable information related to current events (at the time the video has been recorded).

\noindent{\textbf{IV. Persuasion focused}} content includes marketing videos such as \textbf{14. Direct marketing}, which aim to sell products, \textbf{15. Indirect marketing} - aimed to increase brand recognition rather than directly sell a product and \textbf{16. Influencer marketing} content that subtly promotes products. Additionally, \textbf{17. Political propaganda}, \textbf{18. Religious proselytizing}, and \textbf{19. Social movement advocacy} content seek to persuade viewers to support their respective causes. 

\noindent{\textbf{V. Reputation based}} videos pertain to \textbf{20. Character assassination} to damage the reputation of individuals with slander, or \textbf{21. Whistle-blowing}, where the goal is to signal bad practice by subjects or organisations. Other classes include \textbf{22. Life documentation} with individuals sharing their daily life activities. Finally, \textbf{23. Personal-Promotion} content constitutes videos of people describing their work and experience, with the goal of making themselves known or selling their services.

We split the categories in benign and malicious intent-classes as follows. 
\begin{itemize}
    \item \textbf{11 benign categories.} 4. Comedy, 5. Drama / storytelling, 9. Academic or scientific, 10. Guidance-or-Tutorial, 13. News or current event, 14. Direct marketing, 15. Indirect marketing, 19. Social movement advocacy, 21. Whistle-blowing, 22. Life documentation, 23. Personal-Promotion. 
    \item \textbf{12 Malicious categories.} 1. Fake expertise, 2. Financial fraud, 3. Social engineering, 6. Fear-mongering, 7. Outrage generation, 8. Viral sensationalism, 11. Conspiracy theories, 12. Pseudoscience, 16. Influencer marketing, 17. Political propaganda, 18. Religious proselytizing, 20. Character assassination.
\end{itemize}

We ensured the consistency of content across each class through a rigorous manual annotation process.

\begin{algorithm}
\caption{Automatic Dataset Collection}
\begin{algorithmic}[1]
\STATE \textbf{Input:} List of Categories, each with associated Keywords
\STATE \textbf{Output:} Collection of human-centric video links categorized by intent

\FOR{each Category in Categories}
    \FOR{each Keyword in Category}
        \STATE Search YouTube for \textit{Keyword}
        \STATE Select the first 5 videos in the search results
        \STATE Add results to the set of unique videos with the associated Category
    \ENDFOR
\ENDFOR

\STATE Filter out non-human centric videos using Haar cascade

\STATE \textbf{Return} Collection of filtered video links
\end{algorithmic}
\label{alg:data_collec}
\end{algorithm}

\subsection{Dataset Curation}

We curated \textit{keywords} pertaining to each of the $23$ intent classes, by common LLMs (Claude, ChatGPT, Mistral and Meta.ai), filtering them manually, retaining the ones which would be most likely to yield videos of the corresponding class upon searching. Using these keywords as search prompts, we collected 
videos from YouTube, by matching the search terms against video metadata (title, description and tags), and only choosing videos with a duration of less than 4 minutes, in order to enhance variety in the dataset. Roughly 220 videos per class were collected (see Algorithm \ref{alg:data_collec}).

\begin{algorithm}

\caption{Human verification of collected video links}
\begin{algorithmic}[1]
\STATE \textbf{Input:} Collection of (Video, Category) pairs from Step 1
\STATE \textbf{Output:} Final dataset of verified, relevant videos

\FOR{each (Video, Category) in Collection}
    \STATE Human verifies if the video is relevant to the Category
    \IF{video is relevant}
        \STATE Add video to Dataset
    \ELSE
        \STATE Discard video
    \ENDIF
\ENDFOR

\STATE \textbf{Return} Final curated Dataset

\end{algorithmic}
\label{alg:human_verif}
\end{algorithm}

\textbf{Human-centric dataset.} To ensure our dataset contained only human-centric content, we employed Haar cascade classifiers towards detecting faces in video frames, excluding videos that lacked human presence.

\textbf{Manual annotation.} As intent might not be associated to metadata, and therefore keywords, we then perform manual annotation, in order to verify the correspondence of videos and correct intent-categories. For each video, three human annotators are given $3$ options: (i) keep the automated category, (ii) change the category to a different category, (iii) remove the video, in case no category is adequate. The final label was decided by a majority vote among three annotators. Some videos might be ambiguous \textit{w.r.t.} category, \textit{e.g.,} propaganda and fake news. Such videos, causing the human annotators to disagree between the intent labels were discarded from the dataset entirely, retaining only videos with clear intent in a single intent class. While we acknowledge that this might not represent human communication in all its complexity, this is the first work on the subject and future research will investigate more complex labeling for the data. However, we note that using a single label does not degrade the distinction between malicious and benign video as combination of malicious and benign labels would ultimately mean that the video is malicious. After this annotation step, we retain $5168$ high quality videos with accurate class labels. Algorithm \ref{alg:human_verif} shows this human annotation process as a continuation of the video scraping step discussed before. 

\textbf{Transcripts.} Transcripts for videos have been automatically extracted based on timestamps, ensuring alignment between the video content and corresponding textual data for analysis. As a result, at the end of the dataset curation, each video has 3 modalities. 

\subsection{Dataset Statistics}


After the meticulous manual filtering, \textbf{IntentHQ} contains $5168$ high quality, manually verified videos, selected from YouTube. The dataset spans approximately $115$ hours of content. Each of the $23$ predefined intent categories contains between $220$ and $230$ videos, achieving an almost uniform distribution. This corresponds to 2472 videos with benign intent and 2696 videos with malicious intent. The dataset includes visual, audio, and speech transcript modalities, extracted using Whisper-Large \cite{radford2023robust}. 

\subsection{Language and Audio Quality}

The predominant language in IntentHQ is English, which reflects the prevalence of English-language content on YouTube and enhances accessibility for broader research use. Nonetheless, the dataset also includes non-English videos, representing languages such as French, Thai, Hindi, Cantonese, Urdu, and Korean, thereby contributing to linguistic diversity.

While most videos maintain clear and intelligible audio, a small minority exhibits degraded audio quality due to background noise or compression artifacts. However, such instances are limited and do not significantly impact the overall utility of the dataset for multimodal analysis. We perform an ablation, in order to determine the impact of audio to the intent recognition model.

\subsection{Deepfake content}
We did not explicitly exclude deepfake videos during our curation, thus IntentHQ contains both, real and generated videos. However, we note that (a) real videos far outnumber generated videos especially when considering only videos that are entirely generated. (b) In some categories such as \textbf{religious proselytizing}, deepfakes are overrepresented, whereas in others such as \textbf{news or current event} there are no deepfakes present due to the nature of the category.

\section{Baseline Methods}

We implement baseline models by leveraging existing pretrained encoders to extract features independently from each modality: video, audio, and text. These modality-specific features are then fused and passed through a classification network trained to perform intent recognition. This approach allows us to evaluate the effectiveness of feature representations from each modality and their combinations using standard architectures.


\subsection{Simple Multi-Layer Perceptron (MLP)}

In this supervised MLP baseline, features are extracted independently from video, audio, and transcript modalities using frozen encoders. These features are concatenated into a single vector and passed through a shallow feed-forward network (FFN) to predict one of the 23 intent classes. The MLP serves as a simple yet strong classifier to benchmark the quality of the features across different encoder combinations. We elaborate on the employed encoders in the Supplementary Material. 

We experiment with multiple encoder combinations to evaluate their joint representational power. Table~\ref{table:mlp_variants_extended} summarizes the encoder combinations and results for the MLP-based architecture.

\subsection{Simple Cross-Attention}

Towards exploiting modality-specific dependencies, we adopt a second supervised architecture that incorporates cross-attention. In contrast to the FFN-based fusion in the MLP model, this approach utilizes a Transformer decoder architecture to integrate multi-modal representations more effectively.

We begin with the same frozen encoders to extract features for each modality. These features are then projected into a shared embedding space and passed into a stack of Transformer decoder blocks. Within each block, \textit{cross-attention layers} are used, allowing features from one modality (\textit{e.g.,} text) to attend to those from another (\textit{e.g.,} video or audio). This mechanism dynamically weights the importance of each modality's features during classification.

The key intuition is that intent is often implied by interplay between modalities, for instance, sarcasm may be revealed by tonal mismatch between audio and text, or manipulation by subtle gestures in video. Cross-attention allows the model to capture such nuanced inter-modal cues. Table~\ref{table:cross_attention_variants} summarizes the encoder combinations and results for the Cross-Attention-based architecture.

\subsection{Training}
\label{subsec:training}

Feature extraction is performed offline using frozen encoders. For MLP variants, extracted features are concatenated and passed to a 3-layer FFN. For Cross-Attention variants, projected modality-specific embeddings are fed into a Transformer decoder with 4 layers, 8 heads, and residual dropout of $0.2$.

Models are trained using cross-entropy loss
\[
    \mathcal{L} = -\frac{1}{N} \sum_{i=1}^{N} \sum_{c=1}^{C} y_{i,c} \log(\hat{y}_{i,c}),
\]
where $y_{i,c}$ is the one-hot ground truth and $\hat{y}_{i,c}$ is the predicted softmax probability.

\subsection{Hyperparameters}

All models are trained for $30$ epochs with a batch size of $4$, using the Adam optimizer with an initial learning rate of $1 \times 10^{-4}$ and weight decay of $1 \times 10^{-5}$. Learning rate is halved every 3 epochs with no improvement in validation loss. Gradient clipping is applied at 1.0.

We use a 5-fold cross-validation setup, ensuring stratification across intent categories. Results are reported as the mean of Accuracy and F1-Score across folds. For each fold, the best model (lowest validation loss) is retained for evaluation.

\section{Proposed Methodology}

We here present our proposed methodology for intent recognition, which builds upon self-supervised cross-modal alignment followed by supervised fine-tuning for classification. Our objective is to jointly align features from the three modalities: video, audio, and text, through self-supervision, and then use the aligned encoders for downstream intent classification.

\subsection{Dataset Augmentation}

To facilitate self-supervised training at scale, a significant expansion of the dataset was required. Therefore, we implemented an augmentation strategy specifically designed for the intent recognition task.

The key insight behind our augmentation approach is that the \textit{intent} of a video is preserved under semantic paraphrasing of its spoken content. Based on this, we generate \textbf{three paraphrased variants} of each subtitle transcript using the large language model \textit{Meta Llama 3 8b}. Each prompt instructs the model to retain the core meaning and, crucially, the underlying intent of the original text, while varying its linguistic form. This yielded diverse, as well as semantically aligned subtitle variants.

Next, \textbf{pseudo audio} is synthesized from each paraphrased text using text-to-speech (TTS). These synthetic audio clips are then temporally aligned with the original video frames. The result is a set of \textbf{three augmented video instances per original video}, each consisting of (i) the original visual stream, (ii) a paraphrased transcript, and (iii) corresponding pseudo audio.

This multimodal augmentation strategy effectively increases the size of the training set by a factor of four, while focusing on intent recognition. 
The augmented dataset is used exclusively for training the self-supervised model, which is described in the next subsection. We note that evaluation is performed on manually verified real videos from the original dataset.

\subsection{Three-Way Contrastive Self-Supervision on IntentHQ}

We design a self-supervised framework that jointly aligns the three modalities text, audio, and video, via a contrastive learning objective, as shown in Figure \ref{fig:three-way-contrastive}. Using the IntentHQ dataset, which contains approximately 5k videos, we run the self-supervised training of this model.

Our pretraining model employs frozen modality-specific encoders, selected based on their performance among supervised variants. Specifically, we use the CLIP ViT-L/14 encoder for visual frames, WavLM for audio segments, and the CLIP Text Encoder for textual transcripts. The output embeddings from each encoder are projected into a shared latent space using modality-specific linear projection heads.

Towards learning a unified multi-modal representation, we optimize a \textbf{three-way contrastive loss} that encourages alignment between each pair of modalities: text and video, video and audio, and audio and text. For example, the model is trained to maximize similarity between the paraphrased transcript and corresponding visual features, between the visual features and the aligned audio, and between the audio and the transcript features. This tri-directional alignment helps the model learn consistent and semantically meaningful representations across modalities without relying on explicit intent labels.

Positive pairs are constructed from features originating from the same video sample (\textit{e.g.,} a transcript and the audio synthesized from it), while negative pairs are drawn from different samples within the same training batch.

\textbf{Loss Function.} The total training loss is the sum of three pairwise contrastive losses
\[
\mathcal{L}_{\text{total}} = \mathcal{L}_{TV} + \mathcal{L}_{VA} + \mathcal{L}_{AT},
\]
where, $\mathcal{L}_{TV}$ denotes the contrastive loss between text and video, $\mathcal{L}_{VA}$ is the loss between video and audio, and $\mathcal{L}_{AT}$ is the loss between audio and text. Each term is computed using the InfoNCE loss
\[
\mathcal{L}_{ab} = -\log \frac{\exp(\text{sim}(a_i, b_i)/\tau)}{\sum_{j=1}^{N} \exp(\text{sim}(a_i, b_j)/\tau)}.
\]
In this formulation, $a_i$ and $b_i$ are the projected features of modalities $a$ and $b$ for the $i$-th sample in the batch, $\text{sim}(\cdot, \cdot)$ denotes cosine similarity, $N$ is the batch size, and $\tau$ is a learnable or fixed temperature hyperparameter that controls the sharpness of the similarity distribution.

\subsection{Fine-Tuning}

Following pretraining, we freeze the modality encoders and their respective projection heads. A lightweight MLP classifier is then introduced, which takes as input the concatenated embeddings from the three aligned modalities. 
This classifier is fine-tuned to predict one of the 23 intent categories, enabling the model to specialize its general-purpose, cross-modal representation for the downstream intent recognition task. The results of this architecture are reported in Table \ref{tab:ss-results}.

\subsection{Hyperparameters}

\textbf{Pretraining.} For the self-supervised pretraining phase, we trained the model for 50 epochs using the AdamW optimizer with a learning rate of $2 \times 10^{-4}$. The batch size was set to 64 triplets (each consisting of a text, audio, and video instance). We used a contrastive temperature parameter $\tau = 0.07$ to scale the cosine similarities in the InfoNCE loss. Each encoder's output was passed through a two-layer projection head with ReLU activation, mapping embeddings into a shared 768-dimensional latent space. Training was conducted over all three modality pairs, text-video, video-audio, and audio-text simultaneously using a multi-view contrastive objective.

\textbf{Fine-Tuning.} During the fine-tuning stage, all three encoders (CLIP ViT-L/14, WavLM, and CLIP Text) were kept frozen, and a lightweight MLP classifier was trained on top of the concatenated projected features. We trained this classifier for 30 epochs with a batch size of 4, using the AdamW optimizer and a learning rate of $1 \times 10^{-4}$ with a cosine annealing learning rate schedule. Dropout with a rate of 0.3 was applied in the classifier to prevent overfitting. Model performance was assessed using 5-fold cross-validation, and the best-performing model checkpoint was selected based on the lowest validation loss.

\section{Experimental Results and Analysis} 
\label{results}

We present in Table \ref{table:mlp_variants_extended} and Table \ref{table:cross_attention_variants} the results of the model variants evaluated on the IntentHQ dataset using pretrained encoder feature fusion. The overall classification accuracy across all variants 
constitutes $40.3\%$ for 23 classes. Among all, Variants MLP-1 and CA-9 outperform others, achieving accuracies of $40.3\%$ and $44.3\%$, respectively.

\begin{table*}[h]
    \centering
    \resizebox{0.8\linewidth}{!}{
    \begin{tabular}{ccccc}
    \hline
        \textbf{Variant} & \textbf{Video Encoder} & \textbf{Audio Encoder} & \textbf{Text Encoder} & \textbf{Accuracy / F1-Score} \\
    \hline
        \textbf{MLP-1} & CLIP ViT-L/14 & WavLM & LLaMA 3.2 1B & \textbf{40.3\% / 0.386} \\
        MLP-2 & VideoLLaMA3 & WavLM & LLaMA 3.2 1B & 30.6\% / 0.265 \\
        MLP-3 & VifiCLIP & WavLM & LLaMA 3.2 1B & 39.6\% / 0.370 \\
        MLP-4 & ONE-PEACE & ONE-PEACE & LLaMA 3.2 1B & 38.5\% / 0.358 \\
        MLP-5 & CLIP ViT-L/14 & WavLM & CLIP Text Encoder &  39.8\% / 0.379 \\
        MLP-6 & CLIP ViT-B/32 & HuBERT & Qwen 2.5 1.5B & 38.9\% / 0.340 \\
        MLP-7 & SigLIP (NaViT) & Wav2Vec 2.0 & LLaMA 3.2 1B & 37.5\% / 0.325 \\
        MLP-8 & ViT-B/32 & WavLM & BERT-Large & 36.7\% / 0.312 \\
        MLP-9 & VideoMAEv2 & Data2Vec & Qwen 2.5 1.5B & 39.0\% / 0.355 \\
        \textbf{Proposed} & CLIP ViT-L/14 & WavLM & CLIP Text Encoder & \textbf{52.5\% / 0.445} \\
    \hline
    \end{tabular}}
    \caption{Results from Supervised MLP-based classification using different encoder combinations. Accuracy and F1-score are based on 5-fold cross-validation.}
    \label{table:mlp_variants_extended}
\end{table*}

This performance difference is attributed to CLIP ViT-L/14 being employed as video encoder in Variant MLP-1. The CLIP model's robust pretraining on a large-scale image-text corpus enables it to capture semantically rich features that generalize well across diverse intent scenarios. In contrast, Variant MLP-2 employs SigLIP, which, despite being designed for large-batch contrastive learning, suffers from representational sparsity due to its reliance on weaker visual backbones and its focus on language-centric fine-tuning in models like VideoLLaMA3.

Variant MLP-3, based on VifiCLIP, was pretrained on Kinetics-400 and is therefore limited to action-centric understanding. This hampers its capacity to interpret high-level social signals and thematic cues essential for intent classification. Similarly, Variant MLP-4, which utilizes ONE-PEACE for vision and audio, may dilute modality-specific learning in favor of shared representation space generality, making it less effective in nuanced visual reasoning.

\begin{table*}[h]
    \centering
    \resizebox{0.8\linewidth}{!}{
    \begin{tabular}{ccccc}
    \hline
        \textbf{Variant} & \textbf{Video Encoder(s)} & \textbf{Audio Encoder} & \textbf{Text Encoder} & \textbf{Accuracy / F1-Score} \\
    \hline
        CA-1 & CLIP ViT-L/14 & WavLM & LLaMA 3.2 1B & 42.7\% / 0.388 \\
        CA-2 & VideoLLaMA3 & WavLM & LLaMA 3.2 1B & 40.4\% / 0.365 \\
        CA-3 & VifiCLIP & WavLM & LLaMA 3.2 1B & 44.1\% / 0.404 \\
        CA-4 & ONE-PEACE & ONE-PEACE & LLaMA 3.2 1B & 43.0\% / 0.391 \\
        CA-5 & CLIP ViT-L/14 & WavLM & CLIP Text Encoder &  41.8\% / 0.374 \\
        CA-6 & CLIP ViT-B/32 & HuBERT & Qwen 2.5 1.5B & 41.2\% / 0.368 \\
        CA-7 & SigLIP (NaViT) & Wav2Vec 2.0 & LLaMA 3.2 1B & 39.6\% / 0.353 \\
        CA-8 & ViT-B/32 & WavLM & BERT-Large & 38.7\% / 0.341 \\
        \textbf{CA-9} & VideoMAEv2 & Data2Vec & Qwen 2.5 1.5B & \textbf{44.3\% / 0.410} \\
        \textbf{Proposed} & CLIP ViT-L/14 & WavLM & CLIP Text Encoder & \textbf{52.5\% / 0.445} \\
    \hline
    \end{tabular}}
    \caption{Results from Supervised Cross-Attention architecture using different encoder combinations. Accuracy and F1-score are based on 5-fold cross-validation.}
    \label{table:cross_attention_variants}
\end{table*}

The results from the cross-attention variants, presented in Table \ref{table:cross_attention_variants} highlight the superior performance of the CA-9 model, which achieves an accuracy of $44.3\%$ and an F1-score of $0.410$. The performance of the cross-attention-based variants surpasses that of their MLP counterparts in most cases, with CA-9 outperforming the highest MLP variant (MLP-1) by $4.0\%$ in accuracy. This improvement can be attributed to the use of cross-attention mechanisms, which allow the model to better align and integrate information from different modalities, leading to more robust representations. In particular, the combination of VideoMAEv2, Data2Vec, and the Qwen text encoder in CA-9 helps the model to attend to relevant features across modalities with greater precision, enabling more effective intent classification.

Table I in the Supplementary Material presents the performance of the models after pretraining on large-scale public datasets, followed by supervised fine-tuning on IntentHQ. Notably, Kinetics, VGGSound, and SSv2 pretraining datasets yield relatively low accuracy, precision, and recall, indicating that these datasets may not provide the necessary diversity and complexity to capture the wide range of intents in the IntentHQ dataset. While these results reflect some improvement, they fall short of the performance achieved with the self-supervised pretraining approach, further emphasizing the need for better-suited training strategies.

Results in Table \ref{tab:ss-results} suggest that using self-supervision vastly improves performance and yields the \textbf{best accuracy across all models}: $52.5\%$. Therefore, our proposed method represents currently the state of the art for intent recognition. This substantial improvement highlights the impact of self-supervised learning in learning rich representations from large amounts of unlabelled data.  Therefore our proposed method is able to learn semantically meaningful features across multiple modalities, allowing for recognition of fine-grained intent classes.

\begin{table}[h]
    \centering
    \resizebox{0.9\linewidth}{!}{
    \begin{tabular}{lc}
    \hline
        \textbf{Model} & \textbf{Accuracy / F1-Score} \\
    \hline
        
        VideoMAE & 35.2\% / 0.291 \\
        ViT-B/16 & 25.4\% / 0.204 \\
        TimeSFormer & 39.8\% / 0.335 \\
        ViViT & 41.2\% / 0.362 \\
        \textbf{Proposed (Ours)} & \textbf{52.5\% / 0.445} \\
    \hline
    \end{tabular}}
    \caption{Comparison of our proposed multimodal intention-recognition model against standard video classification baselines, trained and tested against the augmented dataset}
    \label{table:video_baselines}
\end{table}

In Table \ref{table:video_baselines} we compare our approach to prominent methods in video classification, namely VideoMAE \cite{wang2023videomaev2}, ViT-B/16, TimeSFormer \cite{gberta_2021_ICML}, ViViT \cite{9710415}. We observe that the proposed approach outperforms other methods by at least 10\%.

Alternatively, if we consider the task of binary classification between malicious and benign intent, then our approach achieves an accuracy of 75.56\%.

While an accuracy of 52.5\% might seem low, the chance level for 23 classes would be 4.3\%, which is significantly lower. Meanwhile video classification achieves at best 40\% accuracy, showcasing the need for multi-modal intent recognition methods and how challenging intent recognition, a task on which even humans might be challenged, can be. We observe that confusion occurs mostly between classes that can be difficult to differentiate for humans too, \textit{i.e.,} whistle blowing and conspiracy theory, which can have the same presentation but change only in the veracity of the fact they present, or  pseudo-science and fake expertise, which can overlap. In contrast, intent classes such as financial fraud or religious proselytizing are easier to recognize due to presence of elements (visual or semantic) specific to those classes (\textit{e.g.,} religious symbols).
While multi-label classification could benefit from overlapping labels, it would not be instrumental for the kind of problem encountered with whistle blowing versus conspiracy theory. Here, both classes share similarities and present their facts as true. Only an understanding of broader context, outside of the content of the video, can provide a reliable classification. This might require the use of powerful LLM, however even then the differentiation might be challenging. We note that even humans are challenged, being misguided by conspiracy theories or on the contrary rejecting whistle-blower as fake.

\begin{table}[h!]
\centering
\resizebox{\linewidth}{!}{
\begin{tabular}{lccc}
\hline
\textbf{Training Type} & \textbf{Accuracy} & \textbf{Precision} & \textbf{Recall} \\
\hline
Supervised only with MLP & 40.3\% & 0.38 & 0.36 \\
Supervised only with Cross-Attention & 44.3\% & 0.42 & 0.40 \\
\textbf{Self-supervised + fine-tuning (Proposed)} & \textbf{52.5\%} & \textbf{0.45} & \textbf{0.44} \\
\hline
\end{tabular}}
\caption{Effect of self-supervised pretraining on classification performance.}
\label{tab:ss-results}
\end{table}

\subsection{Contribution of Each Modality}
\label{subsec:ablation}

Towards establishing the contribution of each modality, we conduct an ablation study on Variant MLP-1, the best-performing MLP-based model. As reported in Table \ref{table:ablation}, we compare classification performance when using single, dual, and all three modalities. We find that solely the video modality achieves $37.9\%$ accuracy, indicating its dominance in the task. This is consistent with the observation that visual elements, such as scene composition, gestures, and speaker framing are strong indicators of communicative intent.

Textual information has a high impact, achieving $37.6\%$ accuracy on its own, highlighting the value of linguistic structure and phrasing in discerning intent. Audio performs significantly worse in isolation ($11.7\%$), possibly due to the presence of background music, inconsistent speech quality, and noisy recording conditions.

Interestingly, fusing modalities leads to improved performance only when the modalities carry complementary information. For instance, combining video and text (Combination v) yields the highest performance at $42.2\%$, whereas adding audio to this pair (Combination vii) slightly decreases accuracy to $40.3\%$. This suggests that there is room for improvement \textit{w.r.t.}  fusion. 

\begin{table*}[h]
    \centering
    \resizebox{0.8\linewidth}{!}{
    \begin{tabular}{c|ccc | cccc}
    \hline
        \textbf{Combination} & \textbf{Video} & \textbf{Audio} & \textbf{Text} & \textbf{Accuracy} & \textbf{Precision} & \textbf{Recall} &\textbf{F1-score} \\
    \hline
        i & \ding{51} & \ding{55} & \ding{55} & 37.9\% & 37.0\% & 37.9\% & 0.364 \\
        ii & \ding{55} & \ding{51} & \ding{55} & 11.7\% & 10.9\% & 11.7\% & 0.108 \\
        iii & \ding{55} & \ding{55} & \ding{51} & 37.6\% & 38.2\% & 37.6\% & 0.364 \\
        iv & \ding{51} & \ding{51} & \ding{55} & 38.6\% & 38.6\% & 38.6\% & 0.372 \\
        v & \ding{51} & \ding{55} & \ding{51} & \textbf{42.2\%} & \textbf{41.6\%} & \textbf{42.2\%} & \textbf{0.406} \\
        vi & \ding{55} & \ding{51} & \ding{51} & 33.9\% & 34.0\% & 33.9\% & 0.325 \\
        vii & \ding{51} & \ding{51} & \ding{51} & 40.3\% & 39.5\% & 40.3\% & 0.386 \\
    \hline
    \end{tabular}}
    \caption{Ablation study on impact of the modalities, video, audio, and text on the classification accuracy.}
    \label{table:ablation}
\end{table*}

\subsection{Filtering Out Non-English Videos Improves Performance}

To study the effect of language, we filter the dataset to include only English-language videos. This subset removes multilingual content that may suffer from translation ambiguity or transcript misalignment. As shown in Table~\ref{tab:audio-quality}, classification accuracy on the filtered subset improves by nearly 3 percentage points, confirming that linguistic consistency enhances model reliability. Non-English transcripts, especially when machine-generated, introduce noise that adversely affects both, text and audio alignment.

\subsection{Filtering Out Poor Audio Quality Improves Performance}

Similarly, we assess the role of audio clarity by excluding videos with significant background noise, echo, or distortion. Table~\ref{tab:audio-quality} presents the results before and after filtering. The improvement is consistent - about 1.5 percentage points in accuracy. This suggests that clearer audio contributes to improved alignment and less noise during training, particularly when audio is combined with transcript and visual data.

\begin{table}[h!]
\centering
\begin{tabular}{lccc}
\hline
\textbf{Dataset} & \textbf{Accuracy} & \textbf{Precision} & \textbf{Recall} \\
\hline
Original Dataset & 40.3\% & 39.5\% & 40.3\% \\
Filtered Dataset (English Only) & \textbf{43.0\%} & \textbf{42.1\%} & \textbf{43.0\%} \\
Filtered Dataset (Clear Audio Only) & \textbf{41.8\%} & \textbf{40.9\%} & \textbf{41.8\%} \\
\hline
\end{tabular}
\caption{Effect of audio-quality and language-based filtering on model performance.}
\label{tab:audio-quality}
\end{table}

\subsection{Category Based Performance Disparity}

Confusion matrices for selected combinations are shown in the Supplementary Material. Categories I. Deception and V. Reputation are particularly vulnerable to misclassification, while Category IV. Persuasion, which includes highly structured video types such as marketing and propaganda, shows the best accuracy. This again underscores the role of visual regularities and overt narrative cues in supporting intent recognition.

The confusion matrices for each variant are presented in the Supplementary Material, Figure 1. An inspection of the confusion matrices reveals a consistent disparity between well-performing and poorly performing intent classes. Notably, the categories of \textbf{Conspiracy Theory}, \textbf{Influencer Marketing}, and \textbf{Religious Propaganda} are consistently well-classified. For the former two, the vocabulary and visual symbolism used in such videos tend to be highly specific and recognizable, while for the latter, the spatial composition (a single speaker in a home setting) provides strong visual priors.

On the contrary, categories such as \textbf{Social Engineering} and \textbf{Character Assassination} perform poorly. This is likely due to the subtle, context-dependent cues required to interpret such videos, where visual and verbal manipulation may be concealed under seemingly neutral delivery. These results reflect the limitations of current video encoders in modeling affective and intentional subtext.

These patterns reflect the challenges inherent in subjective or context-heavy intent classes. 

\subsection{Video duration}
In this work we focus on classifying video that are shorter than 4 minutes. While this is sufficient for short format videos (such as the one found on TikTok or X), this excludes many videos that adopt a longer format. 
Future work will investigate the classification of longer video and include long video in future datasets.

\section{Conclusions}


In this work, we propose the novel task of \textbf{intent recognition in videos}, which goes beyond traditional video understanding by analyzing the underlying goal or motivation of a video, such as persuasion, misinformation, or personal promotion, rather than simply identifying surface-level content or actions. Given the increasing prevalence of both, real and AI-generated content that can influence public opinion, consumer behavior, and sociopolitical discourse, this task carries high societal impact.

To advance research in this direction, we introduce \textit{IntentHQ}, a new dataset comprising $5,168$ high-quality human-centric videos, incorporating video, audio, and transcript. We manually annotate IntentHQ into \textit{23 fine-grained intent classes}, organized under five overarching categories. 

We benchmark intent classification on IntentHQ using \textit{multiple state-of-the-art multimodal architectures}, each combining frozen pre-trained encoders for video, audio, and text with a supervised classifier. The best-performing configuration employs \textit{CLIP ViT-L/14 for video}, \textit{WavLM for audio}, and \textit{LLaMA 3.2 1B for text}, achieving a top-1 accuracy of \textit{40.3\%} on 5-fold cross-validation. An extensive ablation study reveals that while all three modalities contribute useful signals, video features remain the most predictive.

We also introduce a \textit{three-way contrastive self-supervised learning framework}, in order to align the latent spaces of all three modalities video, audio, and text, using augmented pairs. This pretraining method significantly improves downstream classification performance, with accuracy increasing to \textbf{52.5\%} upon fine-tuning. 

Further analyses reveal consistent challenges in recognizing subtle, socially embedded intents such as \textit{Social engineering} or \textit{Character assassination}, and stronger performance on more structurally consistent classes including \textit{Influencer marketing} and \textit{Conspiracy theory}. We also observe the impact of language and audio quality biases and demonstrate that model performance improves when filtering for English-only and high-quality audio data.

Future work will focus on expanding and diversifying the dataset, refining intent categories, incorporating synthetic content, and exploring dynamic, multi-label, and psychologically grounded models to enable fine-grained 
intent classification.

\subsection{Ethical statement}
IntentHQ contains YouTube videos and associated labels, including sensitive categories such as "fraud", "fake news" or "propaganda"  These labels are provided solely for research purposes, they do not represent factual determinations, legal conclusions, or editorial opinions about the content, creators, or subjects of the videos. They are annotations intended for technical experimentation only. Researchers must ensure that the dataset is used in compliance with ethical standards, applicable laws, and platform policies. It should not be employed to stigmatize individuals, spread misinformation, or make definitive claims about the veracity of specific content. Labels may be subjective or context‑dependent, they should not be interpreted as authoritative or universally accurate.



\ifCLASSOPTIONcaptionsoff
  \newpage
\fi



%

\bibliographystyle{IEEEtran}
\bibliography{IEEEabrv,Bibliography}

@String(CVPR= {IEEE Conf. Comput. Vis. Pattern Recog.})

@String(ICCV= {Int. Conf. Comput. Vis.})

@String(ECCV= {Eur. Conf. Comput. Vis.})

@String(ICLR = {Int. Conf. Learn. Represent.})

@String(AAAI = {AAAI})

@String(CVPR  = {CVPR})

@String(ICCV  = {ICCV})

@String(ECCV  = {ECCV})

@String(ICLR  = {ICLR})

@inproceedings{porcile2024finding,
  title={Finding AI-Generated Faces in the Wild},
  author={Porcile, Gonzalo J Aniano and Gindi, Jack and Mundra, Shivansh and Verbus, James R and Farid, Hany},
  booktitle={Proceedings of CVPR},
  pages={4297--4305},
  year={2024}
}

@article{lin2024detecting,
  title={Detecting Multimedia Generated by Large AI Models: A Survey},
  author={Li Lin and Neeraj Gupta and Yue Zhang and Hainan Ren and Chun-Hao Liu and Feng Ding and Xin Wang and Xin Li and Luisa Verdoliva and Shu Hu},
  journal={ArXiv},
  year={2024},
  volume={abs/2402.00045},
  url={https://api.semanticscholar.org/CorpusID:267365030}
}

@article{moreira2024synthetic,
  title={Synthetic Realities and Artificial Intelligence-Generated Contents},
  author={Moreira, Daniel and Marcel, S{\'e}bastien and Rocha, Anderson},
  journal={IEEE Security \& Privacy},
  volume={22},
  number={3},
  pages={7--10},
  year={2024},
  publisher={IEEE}
}

@article{chaudhary2024large,
  title={Large Language Models as Instruments of Power: New Regimes of Autonomous Manipulation and Control},
  author={Chaudhary, Yaqub and Penn, Jonnie},
  journal={arXiv preprint arXiv:2405.03813},
  year={2024}
}

@article{meinke2024frontier,
  title={Frontier Models are Capable of In-context Scheming},
  author={Meinke, Alexander and Schoen, Bronson and Scheurer, J{\'e}r{\'e}my and Balesni, Mikita and Shah, Rusheb and Hobbhahn, Marius},
  journal={arXiv preprint arXiv:2412.04984},
  year={2024}
}

@InProceedings{pmlr-v139-radford21a,
  title = 	 {Learning Transferable Visual Models From Natural Language Supervision},
  author =       {Radford, Alec and Kim, Jong Wook and Hallacy, Chris and Ramesh, Aditya and Goh, Gabriel and Agarwal, Sandhini and Sastry, Girish and Askell, Amanda and Mishkin, Pamela and Clark, Jack and Krueger, Gretchen and Sutskever, Ilya},
  booktitle = 	 {Proceedings of ICML},
  year = 	 {2021},
}

@misc{llama3modelcard,
  title={Llama 3 Model Card},
  author={AI@Meta},
  year={2024},
  url = {https://github.com/meta-llama/llama3/blob/main/MODEL_CARD.md}
}

@InProceedings{wang2023videomaev2,
    author    = {Wang, Limin and Huang, Bingkun and Zhao, Zhiyu and Tong, Zhan and He, Yinan and Wang, Yi and Wang, Yali and Qiao, Yu},
    title     = {VideoMAE V2: Scaling Video Masked Autoencoders With Dual Masking},
    booktitle = {Proceedings of the IEEE/CVF Conference on Computer Vision and Pattern Recognition (CVPR)},
    month     = {June},
    year      = {2023},
    pages     = {14549-14560}
}

@inproceedings{siddiqui2024dvanet,
  title={DVANet: Disentangling view and action features for multi-view action recognition},
  author={Siddiqui, Nyle and Tirupattur, Praveen and Shah, Mubarak},
  booktitle={Proceedings of AAAI},
  year={2024}
}

@inproceedings{xiong2024modality,
  title={Modality-Collaborative Test-Time Adaptation for Action Recognition},
  author={Xiong, Baochen and Yang, Xiaoshan and Song, Yaguang and Wang, Yaowei and Xu, Changsheng},
  booktitle={Proceedings of CVPR},
  year={2024}
}

@inproceedings{hui2024endow,
  title={Endow sam with keen eyes: Temporal-spatial prompt learning for video camouflaged object detection},
  author={Hui, Wenjun and Zhu, Zhenfeng and Zheng, Shuai and Zhao, Yao},
  booktitle={Proceedings of CVPR},
  year={2024}
}

@inproceedings{mahmud2024ssvod,
  title={SSVOD: Semi-supervised video object detection with sparse annotations},
  author={Mahmud, Tanvir and Liu, Chun-Hao and Yaman, Burhaneddin and Marculescu, Diana},
  booktitle={Proceedings of WACV},
  year={2024}
}

@inproceedings{ristea2024self,
  title={Self-distilled masked auto-encoders are efficient video anomaly detectors},
  author={Ristea, Nicolae-C and Croitoru, Florinel-Alin and Ionescu, Radu Tudor and Popescu, Marius and Khan, Fahad Shahbaz and Shah, Mubarak and others},
  booktitle={Proceedings of CVPR},
  year={2024}
}

@inproceedings{wu2024weakly,
  title={Weakly supervised video anomaly detection and localization with spatio-temporal prompts},
  author={Wu, Peng and Zhou, Xuerong and Pang, Guansong and Yang, Zhiwei and Yan, Qingsen and Wang, Peng and Zhang, Yanning},
  booktitle={Proceedings of ACM MM},
  year={2024}
}

@inproceedings{kim2024you,
  title={Do you remember? dense video captioning with cross-modal memory retrieval},
  author={Kim, Minkuk and Kim, Hyeon Bae and Moon, Jinyoung and Choi, Jinwoo and Kim, Seong Tae},
  booktitle={Proceedings of CVPR},
  year={2024}
}

@inproceedings{zhou2024streaming,
  title={Streaming dense video captioning},
  author={Zhou, Xingyi and Arnab, Anurag and Buch, Shyamal and Yan, Shen and Myers, Austin and Xiong, Xuehan and Nagrani, Arsha and Schmid, Cordelia},
  booktitle={Proceedings of CVPR},
  year={2024}
}

@inproceedings{shao2016slicing,
  title={Slicing convolutional neural network for crowd video understanding},
  author={Shao, Jing and Loy, Chen-Change and Kang, Kai and Wang, Xiaogang},
  booktitle={Proceedings of CVPR},
  year={2016}
}

@inproceedings{wu2019long,
  title={Long-term feature banks for detailed video understanding},
  author={Wu, Chao-Yuan and Feichtenhofer, Christoph and Fan, Haoqi and He, Kaiming and Krahenbuhl, Philipp and Girshick, Ross},
  booktitle={Proceedings of CVPR},
  year={2019}
}

@inproceedings{feichtenhofer2020x3d,
  title={X3d: Expanding architectures for efficient video recognition},
  author={Feichtenhofer, Christoph},
  booktitle={Proceedings of CVPR},
  year={2020}
}

@inproceedings{simonyan2014two,
  title={Two-stream convolutional networks for action recognition in videos},
  author={Simonyan, Karen and Zisserman, Andrew},
  booktitle={Proceedings of NeurIPS},
  year={2014}
}

@inproceedings{wang2016temporal,
  title={Temporal segment networks: Towards good practices for deep action recognition},
  author={Wang, Limin and Xiong, Yuanjun and Wang, Zhe and Qiao, Yu and Lin, Dahua and Tang, Xiaoou and Van Gool, Luc},
  booktitle={Proceedings of ECCV},
  year={2016},
}

@inproceedings{
li2022uniformer,
title={UniFormer: Unified Transformer for Efficient Spatial-Temporal Representation Learning},
author={Kunchang Li and Yali Wang and Gao Peng and Guanglu Song and Yu Liu and Hongsheng Li and Yu Qiao},
booktitle={Proceedings of ICLR},
year={2022},
}

@inproceedings{liu2022video,
  title={Video swin transformer},
  author={Liu, Ze and Ning, Jia and Cao, Yue and Wei, Yixuan and Zhang, Zheng and Lin, Stephen and Hu, Han},
  booktitle={Proceedings of CVPR},
  year={2022}
}

@article{wang2022internvideo,
  title={Internvideo: General video foundation models via generative and discriminative learning},
  author={Wang, Yi and Li, Kunchang and Li, Yizhuo and He, Yinan and Huang, Bingkun and Zhao, Zhiyu and Zhang, Hongjie and Xu, Jilan and Liu, Yi and Wang, Zun and others},
  journal={arXiv preprint arXiv:2212.03191},
  year={2022}
}

@inproceedings{li2024mamba,
  title={Mamba-nd: Selective state space modeling for multi-dimensional data},
  author={Li, Shufan and Singh, Harkanwar and Grover, Aditya},
  booktitle={Proceedings of ECCV},
  year={2024},
}

@inproceedings{park2024videomamba,
  title={Videomamba: Spatio-temporal selective state space model},
  author={Park, Jinyoung and Kim, Hee-Seon and Ko, Kangwook and Kim, Minbeom and Kim, Changick},
  booktitle={Proceedings of ECCV},
  year={2024},
}

@inproceedings{reilly2024just,
  title={Just Add $\pi$! pose induced video transformers for understanding activities of daily living},
  author={Reilly, Dominick and Das, Srijan},
  booktitle={Proceedings of CVPR},
  year={2024}
}

@article{soomro2012ucf101,
  title={UCF101: A dataset of 101 human actions classes from videos in the wild},
  author={Soomro, Khurram and Zamir, Amir Roshan and Shah, Mubarak},
  journal={arXiv preprint arXiv:1212.0402},
  year={2012}
}

@inproceedings{caba2015activitynet,
  title={Activitynet: A large-scale video benchmark for human activity understanding},
  author={Caba Heilbron, Fabian and Escorcia, Victor and Ghanem, Bernard and Carlos Niebles, Juan},
  booktitle={Proceedings of CVPR},
  year={2015}
}

@article{liu2022fineaction,
  title={Fineaction: A fine-grained video dataset for temporal action localization},
  author={Liu, Yi and Wang, Limin and Wang, Yali and Ma, Xiao and Qiao, Yu},
  journal={IEEE Transactions on Image Processing},
  year={2022},
}

@inproceedings{miech2019howto100m,
  title={Howto100m: Learning a text-video embedding by watching hundred million narrated video clips},
  author={Miech, Antoine and Zhukov, Dimitri and Alayrac, Jean-Baptiste and Tapaswi, Makarand and Laptev, Ivan and Sivic, Josef},
  booktitle={Proceedings of ICCV},
  year={2019}
}

@inproceedings{yu2019activitynet,
  title={Activitynet-qa: A dataset for understanding complex web videos via question answering},
  author={Yu, Zhou and Xu, Dejing and Yu, Jun and Yu, Ting and Zhao, Zhou and Zhuang, Yueting and Tao, Dacheng},
  booktitle={Proceedings of AAAI},
  year={2019}
}

@misc{radford2018improving,
  title={Improving language understanding by generative pre-training},
  author={Radford, Alec and Narasimhan, Karthik and Salimans, Tim and Sutskever, Ilya and others},
  year={2018},
}

@article{achiam2023gpt,
  title={Gpt-4 technical report},
  author={Achiam, Josh and Adler, Steven and Agarwal, Sandhini and Ahmad, Lama and Akkaya, Ilge and Aleman, Florencia Leoni and Almeida, Diogo and Altenschmidt, Janko and Altman, Sam and Anadkat, Shyamal and others},
  journal={arXiv preprint arXiv:2303.08774},
  year={2023}
}

@inproceedings{dong2025internlm,
  title={Internlm-xcomposer2-4khd: A pioneering large vision-language model handling resolutions from 336 pixels to 4k hd},
  author={Dong, Xiaoyi and Zhang, Pan and Zang, Yuhang and Cao, Yuhang and Wang, Bin and Ouyang, Linke and Zhang, Songyang and Duan, Haodong and Zhang, Wenwei and Li, Yining and others},
  booktitle={Proceedings of NeurIPS},
  year={2025}
}

@inproceedings{wu2025visionllm,
  title={Visionllm v2: An end-to-end generalist multimodal large language model for hundreds of vision-language tasks},
  author={Wu, Jiannan and Zhong, Muyan and Xing, Sen and Lai, Zeqiang and Liu, Zhaoyang and Chen, Zhe and Wang, Wenhai and Zhu, Xizhou and Lu, Lewei and Lu, Tong and others},
  booktitle={Proceedings of NeurIP},
  year={2025}
}

@inproceedings{ren2024timechat,
  title={Timechat: A time-sensitive multimodal large language model for long video understanding},
  author={Ren, Shuhuai and Yao, Linli and Li, Shicheng and Sun, Xu and Hou, Lu},
  booktitle={Proceedings of CVPR},
  year={2024}
}

@inproceedings{peng2023kosmos,
title={Grounding Multimodal Large Language Models to the World},
author={Zhiliang Peng and Wenhui Wang and Li Dong and Yaru Hao and Shaohan Huang and Shuming Ma and Qixiang Ye and Furu Wei},
booktitle={The Twelfth International Conference on Learning Representations},
year={2024},
}

@article{alayrac2022flamingo,
  title={Flamingo: a visual language model for few-shot learning},
  author={Alayrac, Jean-Baptiste and Donahue, Jeff and Luc, Pauline and Miech, Antoine and Barr, Iain and Hasson, Yana and Lenc, Karel and Mensch, Arthur and Millican, Katherine and Reynolds, Malcolm and others},
  journal={Proceedings of NeurIPS},
  year={2022}
}

@article{yang2024qwen2,
  title={Qwen2. 5 technical report},
  author={Yang, An and Yang, Baosong and Zhang, Beichen and Hui, Binyuan and Zheng, Bo and Yu, Bowen and Li, Chengyuan and Liu, Dayiheng and Huang, Fei and Wei, Haoran and others},
  journal={arXiv preprint arXiv:2412.15115},
  year={2024}
}

@article{hsu2021hubert,
  title={Hubert: Self-supervised speech representation learning by masked prediction of hidden units},
  author={Hsu, Wei-Ning and Bolte, Benjamin and Tsai, Yao-Hung Hubert and Lakhotia, Kushal and Salakhutdinov, Ruslan and Mohamed, Abdelrahman},
  journal={IEEE/ACM Transactions on Audio, Speech, and Language processing},
  year={2021},
}

@inproceedings{rombach2022high,
  title={High-resolution image synthesis with latent diffusion models},
  author={Rombach, Robin and Blattmann, Andreas and Lorenz, Dominik and Esser, Patrick and Ommer, Bj{\"o}rn},
  booktitle={Proceedings of CVPR},
  year={2022}
}

@article{liu2024audioldm,
  title={Audioldm 2: Learning holistic audio generation with self-supervised pretraining},
  author={Liu, Haohe and Yuan, Yi and Liu, Xubo and Mei, Xinhao and Kong, Qiuqiang and Tian, Qiao and Wang, Yuping and Wang, Wenwu and Wang, Yuxuan and Plumbley, Mark D},
  journal={IEEE/ACM Transactions on Audio, Speech, and Language Processing},
  year={2024},
}

@article{liu2023visual,
  title={Visual instruction tuning},
  author={Liu, Haotian and Li, Chunyuan and Wu, Qingyang and Lee, Yong Jae},
  journal={Proceedings of NeurIPS},
  year={2023}
}

@inproceedings{ma2024dolphins,
  title={Dolphins: Multimodal language model for driving},
  author={Ma, Yingzi and Cao, Yulong and Sun, Jiachen and Pavone, Marco and Xiao, Chaowei},
  booktitle={Proceedings of ECCV},
  year={2024}
}

@inproceedings{
  tang2023salmonn,
  title={SALMONN: Towards Generic Hearing Abilities for Large Language Models},
  author={Changli Tang and Wenyi Yu and Guangzhi Sun and Xianzhao Chen and Tian Tan and Wei Li and Lu Lu and Zejun MA and Chao Zhang},
  booktitle={The Twelfth International Conference on Learning Representations},
  year={2024},
}

@inproceedings{xu2024pointllm,
  title={Pointllm: Empowering large language models to understand point clouds},
  author={Xu, Runsen and Wang, Xiaolong and Wang, Tai and Chen, Yilun and Pang, Jiangmiao and Lin, Dahua},
  booktitle={Proceedings of ECCV},
  year={2024}
}

@article{yang2023mm,
  title={Mm-react: Prompting chatgpt for multimodal reasoning and action},
  author={Yang, Zhengyuan and Li, Linjie and Wang, Jianfeng and Lin, Kevin and Azarnasab, Ehsan and Ahmed, Faisal and Liu, Zicheng and Liu, Ce and Zeng, Michael and Wang, Lijuan},
  journal={arXiv preprint arXiv:2303.11381},
  year={2023}
}

@inproceedings{moon2024anymal,
  title={Anymal: An efficient and scalable any-modality augmented language model},
  author={Moon, Seungwhan and Madotto, Andrea and Lin, Zhaojiang and Nagarajan, Tushar and Smith, Matt and Jain, Shashank and Yeh, Chun-Fu and Murugesan, Prakash and Heidari, Peyman and Liu, Yue and others},
  booktitle={Proceedings of EMNLP},
  year={2024}
}

@article{chen2023x,
  title={X-llm: Bootstrapping advanced large language models by treating multi-modalities as foreign languages},
  author={Chen, Feilong and Han, Minglun and Zhao, Haozhi and Zhang, Qingyang and Shi, Jing and Xu, Shuang and Xu, Bo},
  journal={arXiv preprint arXiv:2305.04160},
  year={2023}
}

@inproceedings{lai2024lisa,
  title={Lisa: Reasoning segmentation via large language model},
  author={Lai, Xin and Tian, Zhuotao and Chen, Yukang and Li, Yanwei and Yuan, Yuhui and Liu, Shu and Jia, Jiaya},
  booktitle={Proceedings of CVPR},
  year={2024}
}

@inproceedings{jin2024video,
author = {Jin, Yang and Sun, Zhicheng and Xu, Kun and Chen, Liwei and Jiang, Hao and Huang, Quzhe and Song, Chengru and Liu, Yuliang and Zhang, Di and Song, Yang and Gai, Kun and Mu, Yadong},
title = {Video-LaVIT: unified video-language pre-training with decoupled visual-motional tokenization},
year = {2024},
booktitle = {Proceedings of the 41st International Conference on Machine Learning},
}

@inproceedings{zhang2023speechgpt,
    title = "{S}peech{GPT}: Empowering Large Language Models with Intrinsic Cross-Modal Conversational Abilities",
    author = "Zhang, Dong  and
      Li, Shimin  and
      Zhang, Xin  and
      Zhan, Jun  and
      Wang, Pengyu  and
      Zhou, Yaqian  and
      Qiu, Xipeng",
    booktitle = "Findings of the Association for Computational Linguistics: EMNLP 2023",
    year = "2023"
}

@inproceedings{wu2024next,
  title={Next-gpt: Any-to-any multimodal llm},
  author={Wu, Shengqiong and Fei, Hao and Qu, Leigang and Ji, Wei and Chua, Tat-Seng},
  booktitle={Proceedings of ICML},
  year={2024}
}

@article{wang2024tool,
  title={Tool-lmm: A large multi-modal model for tool agent learning},
  author={Wang, Chenyu and Luo, Weixin and Chen, Qianyu and Mai, Haonan and Guo, Jindi and Dong, Sixun and Li, Zhengxin and Ma, Lin and Gao, Shenghua and others},
  journal={arXiv e-prints},
  pages={arXiv--2401},
  year={2024}
}

@inproceedings{liu2024controlllm,
  title={Controlllm: Augment language models with tools by searching on graphs},
  author={Liu, Zhaoyang and Lai, Zeqiang and Gao, Zhangwei and Cui, Erfei and Li, Ziheng and Zhu, Xizhou and Lu, Lewei and Chen, Qifeng and Qiao, Yu and Dai, Jifeng and others},
  booktitle={Proceedings of ECCV},
  year={2024},
}

@inproceedings{radford2023robust,
  title={Robust speech recognition via large-scale weak supervision},
  author={Radford, Alec and Kim, Jong Wook and Xu, Tao and Brockman, Greg and McLeavey, Christine and Sutskever, Ilya},
  booktitle={Proceedings of ICML},
  pages={28492--28518},
  year={2023},
  organization={PMLR}
}

@article{wang2024deepfake,
  title={Deepfake detection: A comprehensive survey from the reliability perspective},
  author={Wang, Tianyi and Liao, Xin and Chow, Kam Pui and Lin, Xiaodong and Wang, Yinglong},
  journal={ACM Computing Surveys},
  volume={57},
  number={3},
  pages={1--35},
  year={2024},
  publisher={ACM New York, NY}
}

@article{tolosana2020deepfakes,
  title={Deepfakes and beyond: A survey of face manipulation and fake detection},
  author={Tolosana, Ruben and Vera-Rodriguez, Ruben and Fierrez, Julian and Morales, Aythami and Ortega-Garcia, Javier},
  journal={Information Fusion},
  volume={64},
  pages={131--148},
  year={2020},
  publisher={Elsevier}
}

@inproceedings{cozzolino2021spoc,
  title={SpoC: Spoofing camera fingerprints},
  author={Cozzolino, Davide and Thies, Justus and Rossler, Andreas and Nie{\ss}ner, Matthias and Verdoliva, Luisa},
  booktitle={Proceedings of the CVPR},
  pages={990--1000},
  year={2021}
}

@inproceedings{shiohara2022detecting,
  title={Detecting deepfakes with self-blended images},
  author={Shiohara, Kaede and Yamasaki, Toshihiko},
  booktitle={Proceedings of CVPR},
  pages={18720--18729},
  year={2022}
}

@article{zhao2023istvt,
  title={ISTVT: interpretable spatial-temporal video transformer for deepfake detection},
  author={Zhao, Cairong and Wang, Chutian and Hu, Guosheng and Chen, Haonan and Liu, Chun and Tang, Jinhui},
  journal={IEEE Transactions on Information Forensics and Security},
  volume={18},
  pages={1335--1348},
  year={2023},
  publisher={IEEE}
}

@article{kaddar2024deepfake,
  title={Deepfake detection using spatiotemporal transformer},
  author={Kaddar, Bachir and Fezza, Sid Ahmed and Akhtar, Zahid and Hamidouche, Wassim and Hadid, Abdenour and Serra-Sagrist{\`a}, Joan},
  journal={ACM Transactions on Multimedia Computing, Communications and Applications},
  year={2024},
  publisher={ACM New York, NY}
}

@article{talevich2017toward,
  title={Toward a comprehensive taxonomy of human motives},
  author={Talevich, Jennifer R and Read, Stephen J and Walsh, David A and Iyer, Ravi and Chopra, Gurveen},
  journal={PloS one},
  volume={12},
  number={2},
  pages={e0172279},
  year={2017},
  publisher={Public Library of Science San Francisco, CA USA}
}

@article{tur2011spoken,
  title={Spoken language understanding: Systems for extracting semantic information from speech},
  author={Tur, Gokhan and De Mori, Renato},
  journal={IEEE Signal Processing Magazine},
  volume={28},
  number={1},
  pages={16--31},
  year={2011},
  publisher={IEEE}
}

@article{chen2019bertintent,
  title={BERT for joint intent classification and slot filling},
  author={Chen, Qian and Zhuo, Zhu and Wang, Wen},
  journal={arXiv preprint arXiv:1902.10909},
  year={2019},
  publisher={arXiv}
}

@inproceedings{zhang2020task,
  title={Task-Oriented Dialogue with Transferable Multi-Hop Memories},
  author={Zhang, Zhaojiang and Ou, Zihan and Song, Yixuan and Li, Xiaofei and Yu, Zhou},
  booktitle={Proceedings of the 58th Annual Meeting of the Association for Computational Linguistics},
  pages={9182--9192},
  year={2020},
  organization={ACL}
}

@article{liu2019roberta,
  title={RoBERTa: A Robustly Optimized BERT Pretraining Approach},
  author={Liu, Yinhan and Ott, Myle and Goyal, Naman and Du, Jingfei and Joshi, Mandar and Chen, Danqi and Levy, Omer and Lewis, Mike and Zettlemoyer, Luke and Stoyanov, Veselin},
  journal={arXiv preprint arXiv:1907.11692},
  year={2019},
  publisher={arXiv}
}

@inproceedings{casanueva2020efficient,
  title={Efficient intent detection with dual sentence encoders},
  author={Casanueva, I{\~n}igo and Shen, Tong and Coope, Sam and Hoang, Dat and Larsen, Thomas},
  booktitle={Proceedings of the 2nd Workshop on Natural Language Processing for Conversational AI},
  pages={38--45},
  year={2020},
  organization={ACL}
}

@inproceedings{devlin2018bert,
    title = "{BERT}: Pre-training of Deep Bidirectional Transformers for Language Understanding",
    author = "Devlin, Jacob  and
      Chang, Ming-Wei  and
      Lee, Kenton  and
      Toutanova, Kristina",
    booktitle = "Proceedings of the 2019 Conference of the North {A}merican Chapter of the Association for Computational Linguistics: Human Language Technologies",
    year = "2019",
}

@ARTICLE{10477989,
  author={Hashmi, Ehtesham and Yayilgan, Sule Yildirim and Yamin, Muhammad Mudassar and Ali, Subhan and Abomhara, Mohamed},
  journal={IEEE Access}, 
  title={Advancing Fake News Detection: Hybrid Deep Learning With FastText and Explainable AI}, 
  year={2024}}

@ARTICLE{10568915,
  author={Xu, Cheng and Kechadi, M-Tahar},
  journal={IEEE Access}, 
  title={An Enhanced Fake News Detection System With Fuzzy Deep Learning}, 
  year={2024}}

@inproceedings{gberta_2021_ICML,
    author  = {Gedas Bertasius and Heng Wang and Lorenzo Torresani},
    title = {Is Space-Time Attention All You Need for Video Understanding?},
    booktitle   = {Proceedings of the International Conference on Machine Learning (ICML)}, 
    month = {July},
    year = {2021}
}

@INPROCEEDINGS{9710415,

  author={Arnab, Anurag and Dehghani, Mostafa and Heigold, Georg and Sun, Chen and Lučić, Mario and Schmid, Cordelia},

  booktitle={2021 IEEE/CVF International Conference on Computer Vision (ICCV)}, 

  title={ViViT: A Video Vision Transformer}, 

  year={2021}}

%

\vspace{-15mm}
\begin{IEEEbiography}[{\includegraphics[width=1in,height=1.25in,clip,keepaspectratio]{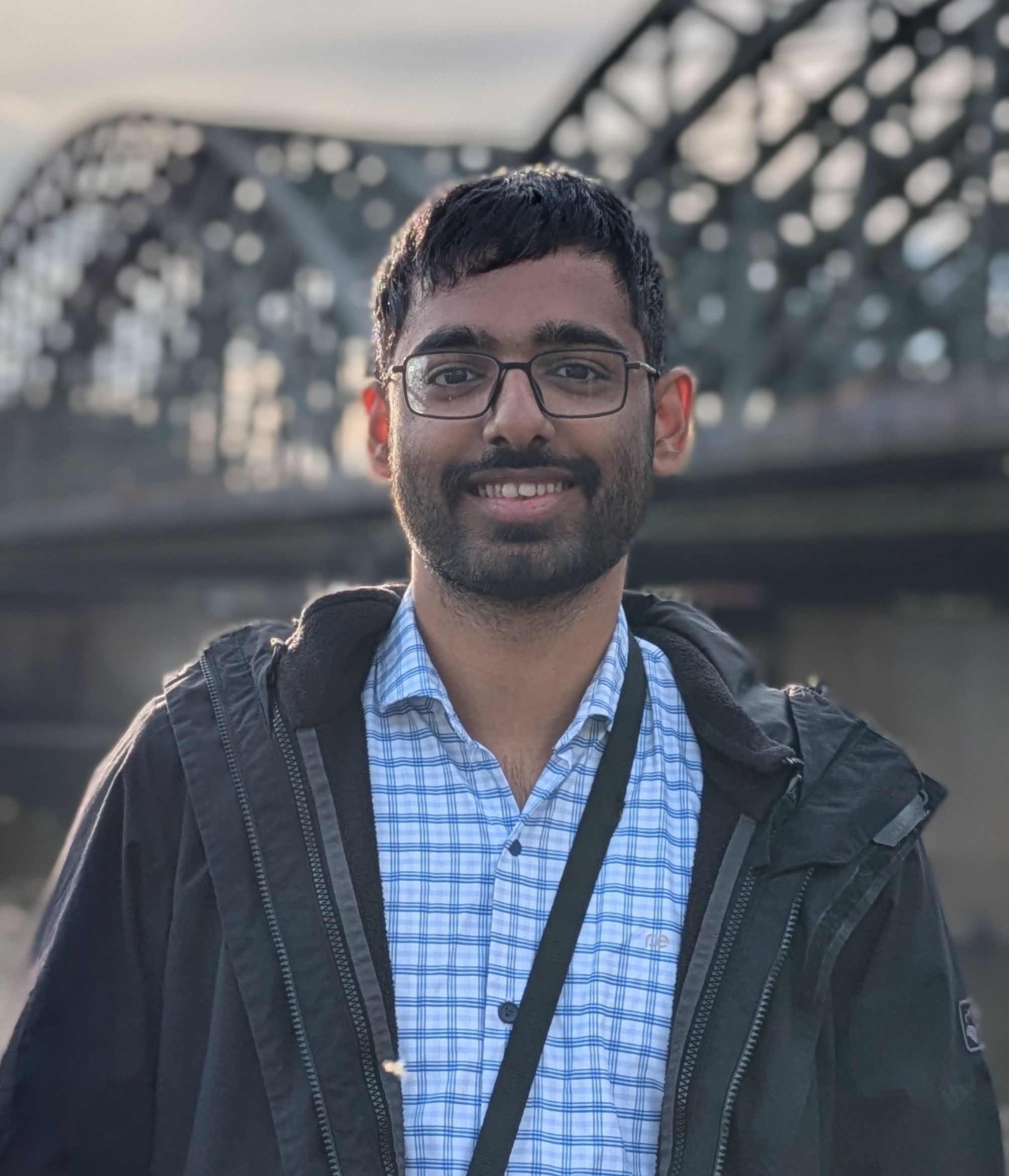}}]{Saurabh Atreya} is currently pursuing a Bachelor of Engineering (B.E.) in Computer Science at BITS Pilani, Hyderabad, India. He is currently a Research Intern at Inria, France. He is a recipient of the Kishore Vaigyanik Protsahan Yojana (KVPY) Fellowship (AIR 110, SA-Stream) awarded by IISc, and the National Talent Search Examination (NTSE) Scholarship awarded by NCERT. 
\end{IEEEbiography}

\vspace{-15mm}
\begin{IEEEbiography}[{\includegraphics[width=1in,height=1.25in,clip,keepaspectratio]{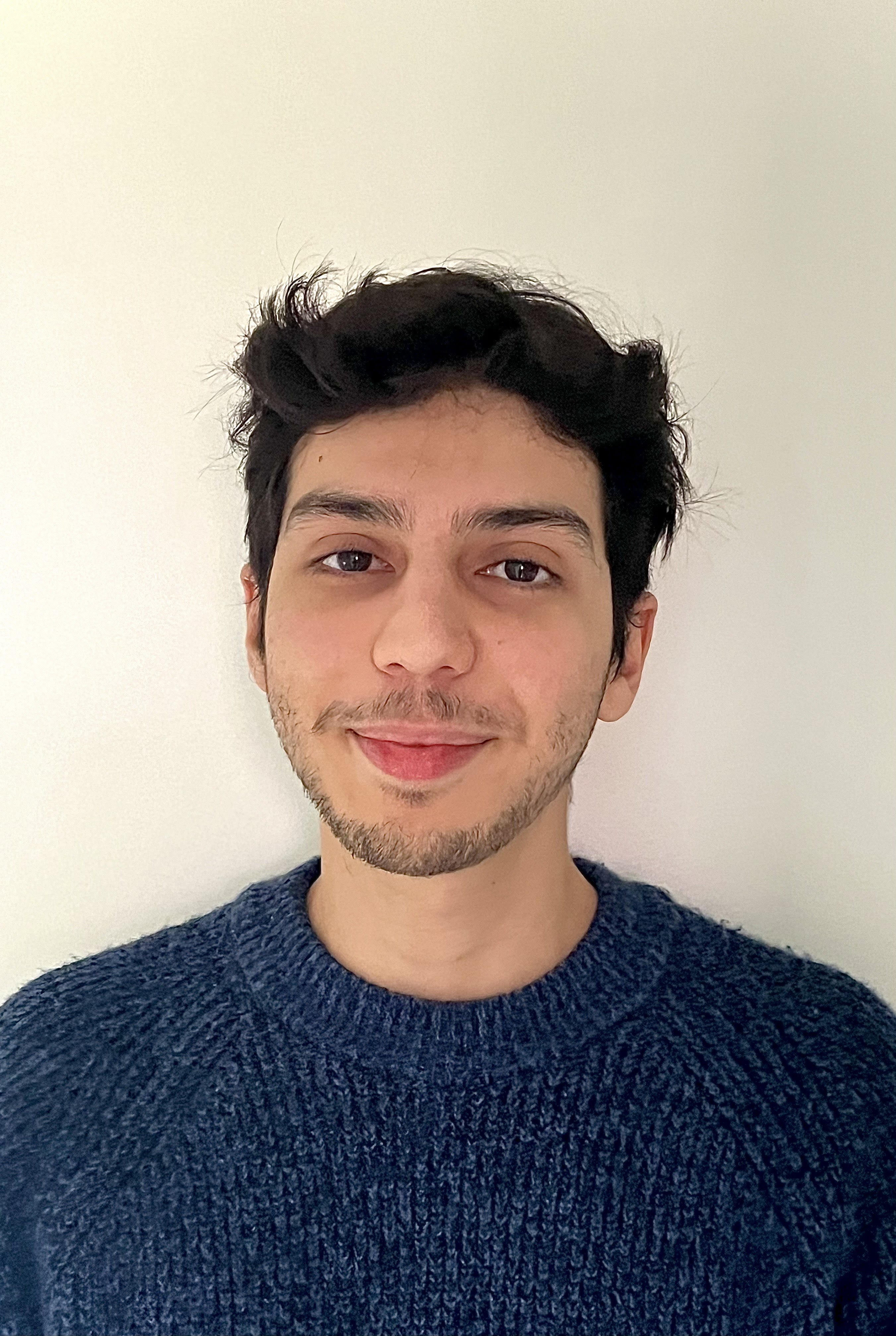}}]{Nabyl Quignon} is a Ph.D. student at Inria under Dr. Antitza Dantcheva's supervision. Previously, he obtained his M.Sc. in computer science from University of Orleans. His current research interest is at the intersection of video generation and representation learning.
\end{IEEEbiography}

\vspace{-15mm}
\begin{IEEEbiography}[{\includegraphics[width=1in,height=1.25in,clip,keepaspectratio]{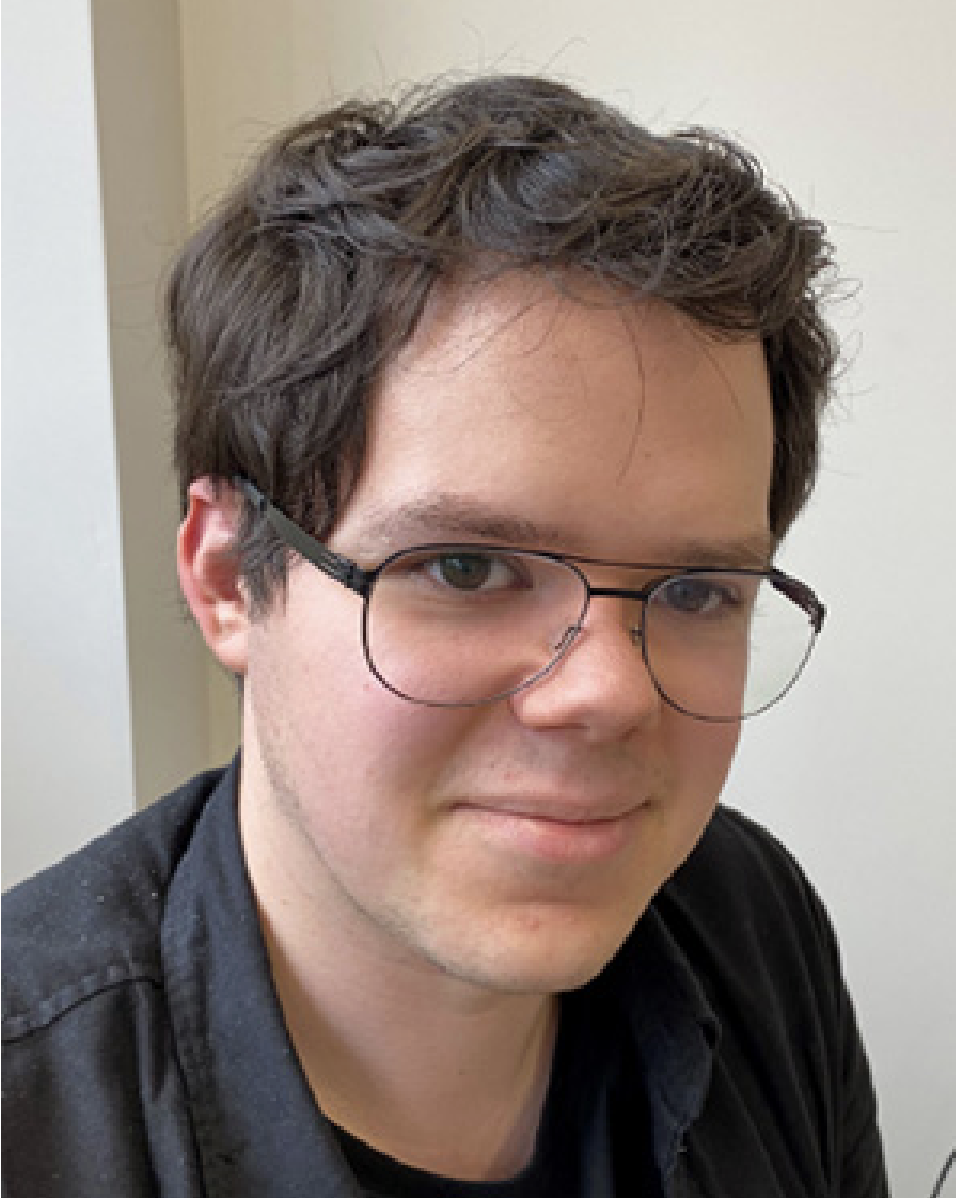}}]{Baptiste Chopin} is currently a Postdoctoral Researcher at Inria, France. Previously, he received his Ph.D. degree at the University of Lille, France and he obtained his engineering degree in computer science from IMT Nord Europe, France. His research concern computer vision and the generation of human motion with application to cognitive sciences.
\end{IEEEbiography}

\vspace{-15mm}
\begin{IEEEbiography}[{\includegraphics[width=1in,height=1.25in,clip,keepaspectratio]{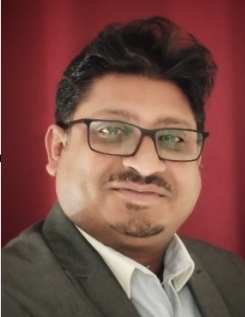}}]{Abhijit Das} is an Assistant Professor at BITS Pilani Hyderabad. Previously, he worked as a Post-Doc Researcher at Inria Sophia Antipolis– Méditerranée, France. He has completed his PhD from the School of Information and Communication Technology, Griffith University, Australia. He is an accomplished machine learning and computer vision researcher with more than 15 years of research and teaching experience. He is presently pursuing an investigation on learning representations and human analysis employing facial and corporeal-based visual features.
\end{IEEEbiography}

\vspace{-15mm}
\begin{IEEEbiography}[{\includegraphics[width=1in,height=1.25in,clip,keepaspectratio]{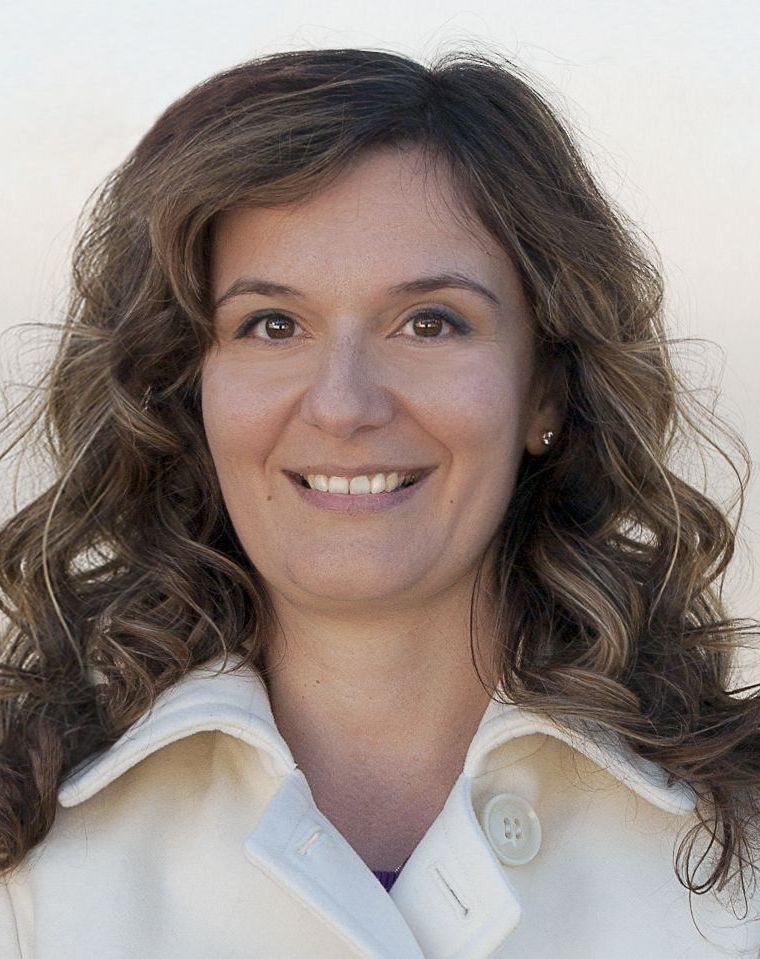}}]{Antitza Dantcheva} Antitza Dantcheva is Directrice de Recherche with the STARS team of Inria Center at Université Côte d'Azur, Sophia Antipolis, France. Previously, she was a Marie Curie fellow at Inria and a Postdoctoral Fellow at the Michigan State University and the West Virginia University, USA. She received her Ph.D. degree from Telecom ParisTech/Eurecom in image processing and biometrics in 2011. Her research is in computer vision and specifically in designing algorithms that seek to learn suitable representations of the human face in interpretation and generation. She is recipient among others of the ANR Jeunes chercheuses / Jeunes chercheurs (JCJC) personal grant, winner of the New Technology Show at ECCV 2022, the Best Poster Award at IEEE FG 2019, winner of the Bias Estimation in Face Analytics (BEFA) Challenge at ECCV 2018 (in the team with Abhijit Das and Francois Bremond) and Best Paper Award (Runner up) at the IEEE International Conference on Identity, Security and Behavior Analysis (ISBA 2017).

\end{IEEEbiography}





\end{document}